\documentclass[journal]{IEEEtran}
\usepackage{cite}
\usepackage{graphicx} 
\usepackage{amsmath}
\usepackage{amssymb}
\usepackage{booktabs}
\usepackage{xcolor}
\usepackage{float}
\usepackage{subcaption} 
\usepackage{afterpage}
\usepackage{subcaption} % For subfigure support
\usepackage{array} % For custom table layouts
\usepackage{amsmath}
\usepackage{algorithmic}
\usepackage{array}
\ifCLASSOPTIONcompsoc
 \usepackage[caption=false,font=normalsize,labelfont=sf,textfont=sf]{subfig}
\else
 \usepackage[caption=false,font=footnotesize]{subfig}
\fi
\usepackage{fixltx2e}
\usepackage{stfloats}
\ifCLASSOPTIONcaptionsoff
 \usepackage[nomarkers]{endfloat}
\let\MYoriglatexcaption\caption
\renewcommand{\caption}[2][\relax]{\MYoriglatexcaption[#2]{#2}}
\fi
\usepackage{url}
\begin{document}
\title{Camouflaged Image Synthesis Is All You Need to Boost Camouflaged Detection}

% \author{Haichao~Zhang,~\IEEEmembership{Student Member,~IEEE,}
%         Yu~Yin, Xinjin~Li,
%         Yun~Fu,~\IEEEmembership{Fellow,~IEEE}
%         % and~Jane~Doe,~\IEEEmembership{Life~Fellow,~IEEE}% <-this % stops a space
% \thanks{Haichao Zhang is PhD Candidate with the Department
% of Electrical and Computer Engineering, Northeastern University, Boston,
% MA, 02115 USA e-mail: zhang.haich@northeastern.edu}% <-this % stops a space
% \thanks{Yu Yin is an Asistant Professor with Case Western Reserve University, Cleveland, OH 44106.}% <-this % stops a space
% \thanks{Xinjin Li is with Columbia University, New York, NY 10027.}% <-this % stops a space
% \thanks{Yun Fu is Professor with Department of the Department
% of Electrical and Computer Engineering \& the Khoury College of Computer Science, Northeastern University, Boston,
% MA, 02115}}

\author{Haichao~Zhang,~\IEEEmembership{Student Member,~IEEE,} 
        Yu~Yin, 
        % Xinjin~Li,
        and Yun~Fu,~\IEEEmembership{Fellow,~IEEE}%
        \thanks{Haichao Zhang is a Ph.D. candidate in the Department of Electrical and Computer Engineering, Northeastern University, Boston, MA 02115, USA. Email: \texttt{zhang.haich@northeastern.edu}}%
        \thanks{Yu Yin is an Assistant Professor in the Department of Electrical Engineering and Computer Science, Case Western Reserve University, Cleveland, OH 44106, USA. Email: \texttt{yu.yin@case.edu}}%
        % \thanks{Xinjin Li is with the Department of Computer Science, Columbia University, New York, NY 10027, USA.}%
        \thanks{Yun Fu is a Professor in the Department of Electrical and Computer Engineering and the Khoury College of Computer Sciences, Northeastern University, Boston, MA 02115, USA. Email: \texttt{yunfu@ece.neu.edu}}}


\markboth{IEEE Transactions on Multimedia, Under Review}%
{Shell \MakeLowercase{Zhang \textit{et al.}}: Camouflaged Image Synthesis}



\maketitle

% \fi
\IEEEpeerreviewmaketitle

\newcommand{\sArt}{state-of-the-art }

\newcommand{\lach}{\vspace{0pt}}
\newcommand{\tickYes}{\bullet}
\newcommand{\cmark}{\ding{51}}
\newcommand{\tickNo}{\hspace{1pt}\ding{55}}
\newcommand{\figref}[1]{Fig.~\ref{#1}}
\newcommand{\tabref}[1]{Table~\ref{#1}}
\newcommand{\eqnref}[1]{(Eq.~\ref{#1})}
\newcommand{\secref}[1]{Section \ref{#1}}

\newcommand{\tabincell}[2]{\begin{tabular}{@{}#1@{}}#2\end{tabular}}

% \newcommand{\AddText}[3]{\put(1,2){\contour{black}{\textbf{\textcolor{green}{3}}}}}
% \newcommand{\trb}[1]{\textbf{\textcolor{black}{#1}}}
% \newcommand{\tgb}[1]{\textcolor{green}{#1}}
% \newcommand{\tbb}[1]{\textcolor{blue}{#1}}
% \newcommand{\supp}[1]{#1}
% \newcommand{\cmm}[1]{\textcolor{blue}{#1}}
% \newcommand{\fdp}[1]{\textbf{\textcolor{red}{#1}}}
% \newcommand{\jgp}[1]{\textcolor{cyan}{#1}}
% \newcommand{}[1]{\textcolor{black}{#1}}

% \maketitle

\begin{abstract}
    % Camouflaged objects that blend into natural scenes pose significant challenges for deep-learning models to detect and synthesize. While camouflaged object detection is a crucial task in computer vision with diverse real-world applications, this research topic has been constrained by limited data availability. We propose a framework for synthesizing camouflage data to enhance the detection of camouflaged objects in natural scenes. Our approach employs a generative model to produce realistic camouflage images, which can be used to train existing object detection models. Specifically, we use a camouflage environment generator supervised by a camouflage distribution classifier to synthesize the camouflage images, which are then fed into our generator to expand the dataset. Our framework outperforms the current state-of-the-art method on three datasets in quantitative and plug-and-play experiments, demonstrating its effectiveness in improving camouflaged object detection. This approach can serve as a plug-and-play data generation and augmentation module for existing camouflaged object detection tasks. 
    % To the best of our knowledge, our task is the first one to propose a deep learning-based camouflage image synthesis task and use it as a data augmentation method.

Camouflaged objects that seamlessly blend into natural scenes pose significant challenges for deep-learning models in both detection and synthesis. Camouflaged object detection is a pivotal task in computer vision with a broad range of real-world applications. However, progress in this area has been hindered by limited access to relevant data. In this paper, we introduce a novel framework for synthesizing camouflage data to enhance the accuracy of camouflaged object detection in natural scenes. Our approach leverages a generative model to produce highly realistic camouflage images, which subsequently bolster the training of existing object detection models. Specifically, we employ a camouflage environment generator guided by a camouflage distribution classifier. This synergy of components facilitates the synthesis of camouflage images, thus augmenting the dataset used for training. The effectiveness of our proposed framework is underscored by its superior performance over the current baselines across three distinct datasets, as demonstrated through quantitative evaluations and plug-and-play experiments. These results validate its prowess in significantly advancing camouflaged object detection accuracy.
Our framework not only outperforms existing methods but also introduces a pioneering deep learning-based camouflage image synthesis task. Furthermore, our approach seamlessly integrates with existing camouflaged object detection tasks, thereby serving as a versatile plug-and-play module for data augmentation. To the best of our knowledge, our work represents the first attempt to propose such a task and effectively employ it as a data augmentation strategy in the realm of camouflaged object detection. We extend current datasets to form a unified camouflage augmented dataset, code and data will be public available.
\end{abstract}

\section{Introduction}
\label{sec:intro}

\begin{figure}[t]
  \centering
  \includegraphics[width=.50\textwidth]{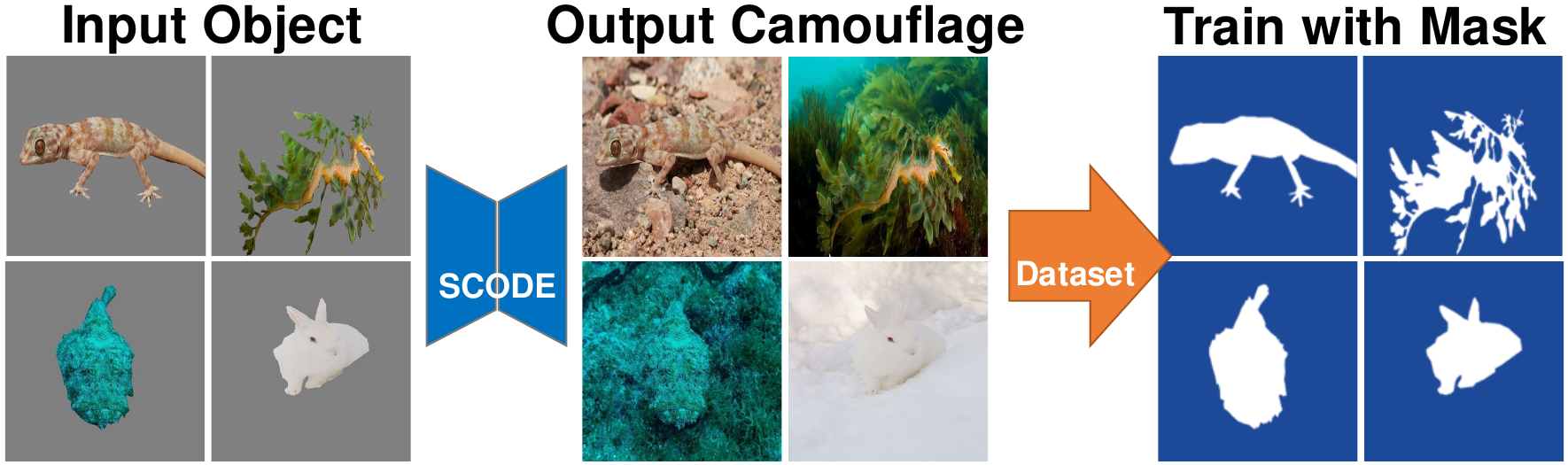}
  \caption{
    The figure shows a set of synthesized camouflage images generated by our Synthetic CamOuflage DEtection (SCODE) model to address the scarcity of data quantity and quality in camouflaged object detection tasks. The gray box represents the input object used by SCODE, the orange box contains generated camouflaged images, and the blue box shows the predicted masks generated by our camouflaged object detection models trained on the synthetic dataset. The synthetic dataset generated by SCODE improves the performance of camouflaged object detection models.
  }  
  % %%\vspace{-15pt}
  \label{fig:head}
\end{figure}

\begin{figure}[t]
  \centering
  \includegraphics[width=0.48\textwidth]{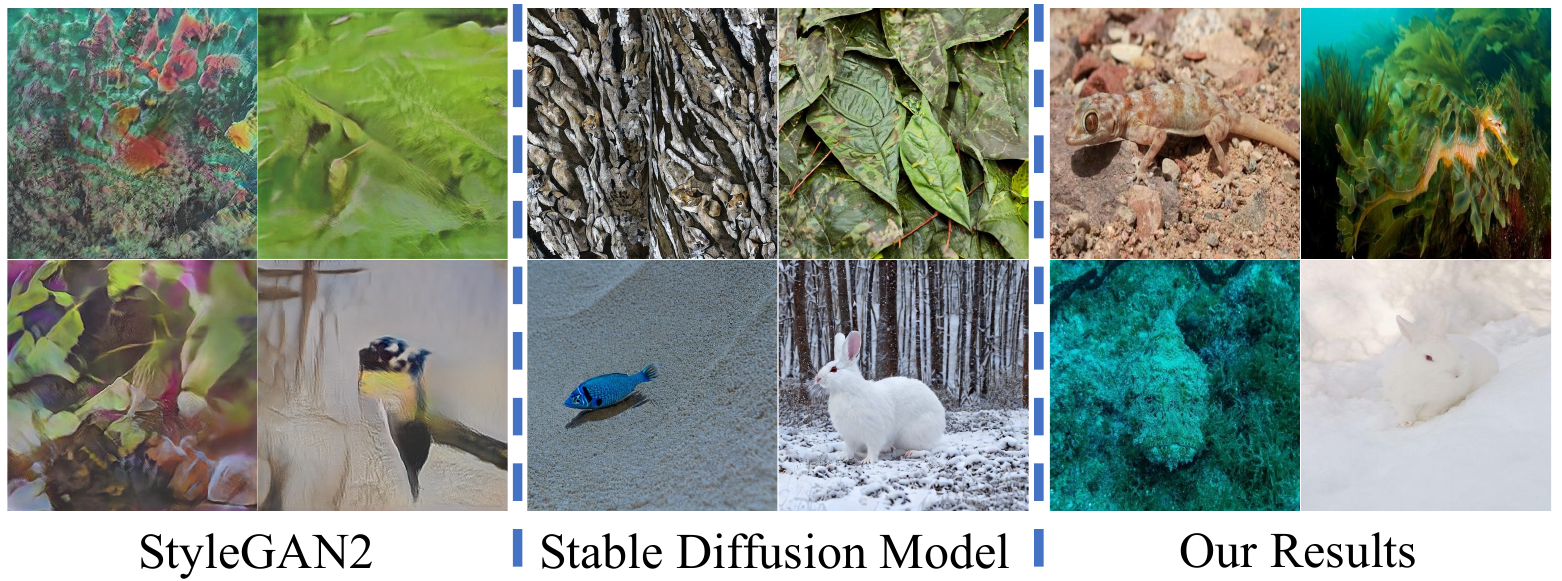}
  %%\vspace*{-15pt}
  \caption{ 
  % \textbf{Qualitative comparison of our model with state-of-the-art generative models.} The four examples shown, from left to right and top to bottom, illustrate various scenarios of camouflage in nature. Our model produces plausible results that successfully synthesize camouflage images. In contrast, the synthesized rabbit and fish in the bottom raw by stable diffusing model~\cite{rombach2021highresolution} do not blend in backgrounds, while the top raw has no foreground objects at all. The results of StyleGAN2~\cite{Karras2019stylegan2} are chaotic and full of artifacts. The details of the experiment setting can be found in Sec.~\ref{subsubsec:figure2}. 
  \textbf{Qualitative comparison of our model with state-of-the-art generative models.} The four examples shown illustrate various scenarios of camouflage in nature, and our model successfully produces plausible results that blend in with the backgrounds. In contrast, the synthesized rabbit and fish in the bottom row by stable diffusing model~\cite{rombach2021highresolution} do not blend in, while the top row has no foreground objects at all. The results of StyleGAN2~\cite{Karras2019stylegan2} exhibit artifacts and are chaotic. %The experiment settings can be found in Sec.~\ref{subsubsec:figure2}.
  \textcolor{red}{\textbf{Please zoom in to see the details.}} 
  }
  \label{fig:headcomparison}
\end{figure}

Deep learning models require large amounts of high-quality data to train effectively, but the scarcity of such data can significantly impact neural network performance, especially for tasks that are difficult to collect and label, such as camouflaged object detection. To overcome this challenge, training machine learning models with synthetic data has emerged as a highly promising direction. Previous works have used computer rendering data~\cite{qi2020learning, girdhar2020cater}, images from video games~\cite{richter2016playing}, or simulations~\cite{li2021solving, wang2020synthesizing} to generate the necessary images for learning. While these approaches significantly advanced the field, the synthetic data they generated still had a gap in realism compared to real-world datasets, limiting the theoretical upper bound of learning performance. Recent advances in generative models, such as generative adversarial networks (GANs)~\cite{goodfellow2020generative}, neural radiance fields (NeRF)~\cite{mildenhall2020nerf}, and diffusion models~\cite{sohl2015deep,rombach2021highresolution}, have dramatically improved the quality of AI-synthesized image generation. Using GAN-synthesized data to train deep learning methods has become an essential choice for specific tasks such as camouflaged object detection. However, currently,  there is less attention paid to camouflaged image synthesis and the camouflaged data is hard to synthesize with current generative models, as shown in  Fig.~\ref{fig:headcomparison}.

Camouflaged object detection (COD)~\cite{fan2020camouflaged} is a task with great potential for medical polyp segmentation, search and rescue systems, and agriculture. The task requires machine learning systems to identify objects hidden and blended in their environments. However, further study of COD is limited by the scarcity of sufficiently large datasets. Although previous works~\cite{chen2025just,chen2025sam,zhang2025learning,pang2024zoomnext,9598866} have collected some datasets~\cite{le2019anabranch,2018Animal,fan2020camouflaged}, the amount of data is still insufficient. Furthermore, increasing the size of the dataset is challenging due to the scarcity of camouflage data, such as the Animal Camouflaged Dataset CHAMELEON~\cite{2018Animal}, which contains only 76 images. These COD data are challenging to capture, record, and label, making them rare and precious. The limited amount of data available for COD currently restricts the performance of existing models.

To address the scarcity of data in camouflaged object detection, one intuitive approach is to collect more data on camouflaged images. However, this is a challenging task since camouflage objects are often rare and difficult to find. Another solution is to use data augmentation methods, but traditional techniques for object detection and segmentation have already been extensively explored, making it necessary to explore other options. Generative methods represent a promising choice for increasing the variety of camouflage objects. For example, Ge et al.~\cite{ge2022dall} use text as an intermediary step to perform data augmentation, describing the components of images with CLIP and then synthesizing the objects and background using DALL-E before randomly placing them on the background.

However, camouflage images pose unique challenges since they are designed to blend in with their surroundings using techniques such as color matching, texture blending, and shape mimicking. Therefore, two significant challenges arise. Firstly, it is difficult for current deep learning models to understand the scene in camouflage images, let alone describe their content. Even segmenting camouflage objects from their background remains a challenging problem that has attracted significant research efforts. Secondly, directly synthesizing camouflage images using current generative models is impractical. While stable diffusion models~\cite{rombach2021highresolution} and StyleGANs~\cite{Karras2019stylegan2} are two of the most promising generative models for image synthesis, stable diffusion models require large amounts of training data, and camouflage data represents only a small fraction of available training data, making it difficult to synthesize accurate camouflage images. StyleGANs, Pix2PixHD~\cite{wang2018pix2pixHD}, and other GAN models cannot recognize the boundaries of camouflage objects, even with the foreground mask as input. As shown in  Fig.~\ref{fig:headcomparison},  the StyleGAN2 model struggles to distinguish between camouflage objects and their background, resulting in chaotic outputs. The Stable Diffusion Model, on the other hand, either fails to synthesize the foreground or generates conspicuous objects that do not blend in with their surroundings. The results from StyleGAN2~\cite{Karras2019stylegan2} and the Stable Diffusion Model~\cite{rombach2021highresolution} show limitations in their ability to generate convincing camouflage images.

To overcome the challenges in camouflaged object synthesis, we propose a method that leverages generative models to synthesize camouflage data and their corresponding segmentation masks, as shown in Fig.~\ref{fig:head}. Since camouflage images involve foreground objects hidden within their surroundings, we start by selecting an image from existing datasets and extracting its foreground object using its mask. We then use our camouflage environment generator (\textbf{CEG}
% , see Sec.~\ref{subsec:ceg}}
) and noise-adding process to generate a realistic background image. However, the synthesized images may not always exhibit the desired camouflage properties. To address this issue, we train a camouflage distribution classifier (\textbf{CDC}
% , see Sec. \ref{subsec:cdc}
) to supervise the synthesis process and ensure that the resulting images are well-camouflaged. Finally, we use the synthesized data to train our camouflaged object detection models. This approach enables us to overcome the challenges of limited data availability and diversity in camouflaged object detection.

% in Sec.\ref{subsec:datasetsynthesis}. 

The contributions of this paper can be listed as follows:
\begin{enumerate}
\item Our method was the first to utilize synthetic data for camouflaged object detection, addressing the scarcity of real-world data. 
% Notably, our work predates a similar approach proposed by ~\cite{he2023strategic} by 6 months.

\item We define and tackle the problem of camouflaged object synthesis. We introduce a camouflage environment generator that synthesizes the background of a given object. This novel approach effectively addresses the challenge of generating realistic and diverse camouflaged images.

\item Our method excels in camouflage data synthesis and surpasses existing methods in the domain of camouflaged object detection. This performance enhancement serves as a testament to the efficacy of synthetic data in elevating model performance.

\end{enumerate}

% The contributions of this paper can be listed as follows,
% \begin{enumerate}
% % %%\vspace{-8pt}
% \item Our method is the first to use synthetic data in camouflaged object detection to address the scarcity of real-world data, 6-months earlier than ~\cite{he2023strategic}.
% % %%\vspace{-8pt}
% \item We define the problem of camouflaged object synthesis and propose a camouflage environment generator to synthesize the background of a given object, which addresses the challenge of generating realistic and diverse camouflage backgrounds.
% % %%\vspace{-8pt}
% \item Our method achieves promising results in camouflage data synthesis and outperforms existing state-of-the-art methods in camouflaged object detection, demonstrating the effectiveness of synthetic data in improving model performance.
% \end{enumerate}

\section{Related Works}
\label{sec:relatedwork}
Our work aims to leverage GAN-synthetic data to address the challenge of limited data availability in camouflaged object detection. We review related works in two areas: Camouflaged object detection and Learning from synthetic data.

\subsection{Camouflaged Object Detection}
\label{subsec:cod}

Camouflaged object detection is gaining significance in computer vision due to its vital role in applications such as military operations, medical imaging, and wildlife monitoring. The first camouflaged object detection method~\cite{tankus1998detection} was introduced in 1998, and more recent approaches have emerged. Rao et al.~\cite{pulla2020camouflaged} proposed a traditional camouflaged object detection method in 2020. Fan et al.~\cite{fan2020camouflaged, fan2021concealed} introduced a camouflaged object detection model and dataset for segmentation.

However, these methods often demand large annotated datasets for robust model training, which poses a challenge for camouflage objects due to their rarity and difficulty in capture. As observed in a recent review~\cite{bi2021rethinking}, camouflage can be categorized into natural creatures using camouflage for survival and military entities employing camouflage to blend into environments during conflicts. Camouflaged object detection seeks to identify concealed targets sharing similarities in texture, color, shape, and size with their surroundings. Up until 2020, only two camouflaged object detection datasets were known. The CHAMELEON~\cite{2018Animal} dataset contains a mere 76 camouflage images without a training set, while the CAMO~\cite{le2019anabranch} dataset comprises only 1250 data points. Despite recent dataset efforts, including COD10K~\cite{fan2020camouflaged}, data scarcity remains a significant hurdle. Our work addresses this challenge by employing GAN-synthetic data for training camouflaged detection.

\subsection{Learning From Synthetic Data}
\label{subsec:syntheticdata}

The quality and quantity of data significantly impact deep learning model performance. When data or labels are challenging to acquire, learning from synthetic data becomes a viable solution. Synthetic data~\cite{zhang2025physrig,zhang2025stable,zhang2024open, zhang2024magicpose4d, zhang2024s3o} has found extensive use in segmentation, autonomous driving, medical images, domain adaptation, 3D reconstruction, and automatic label generation. However, few works directly focus on and perform well in camouflaged image synthesis.

The GTA Dataset~\cite{richter2016playing}, a synthetic dataset from games, has been widely utilized in autonomous driving and computer vision tasks. Qi et al.~\cite{qi2020learning} employed synthetic data to model human-environment relationships. Li et al.~\cite{li2021solving} used synthetic datasets for robotics learning. Acharya et al.~\cite{acharya2021using} harnessed synthetic data as multi-view inputs for multi-view stereo geometry learning. Sankaranarayanan et al.~\cite{sankaranarayanan2018learning} tackled domain shift in semantic segmentation via synthetic data learning. Wang et al.~\cite{wang2019learning} addressed data scarcity in crowd counting by pretraining a model on synthetic datasets. Synthetic data augmentation using GANs has been applied in medical image processing~\cite{frid2018synthetic}.
Recent works like DatasetGAN~\cite{zhang2021datasetgan} utilize StyleGAN~\cite{karras2019style} for realistic image synthesis, generating corresponding semantic segmentation using latent codes. Other studies~\cite{liu2022learning} employ unsupervised domain adaptation for dataset labeling, while ~\cite{mishra2022task2sim} use pre-trained synthetic data models for downstream tasks.

Despite extensive research on synthetic data across various computer vision domains, no prior work has focused on generating synthetic data for camouflage. The distinct challenges of synthesizing camouflage images, including data scarcity, object detection difficulties, foreground-background relationships, and realistic camouflage generation, make this task uniquely challenging. Consequently, the development of a method for synthetic camouflage dataset generation is a pressing research direction for advancing camouflaged object detection.

\begin{figure*}[t]
\centering
\includegraphics[width=1.\linewidth]{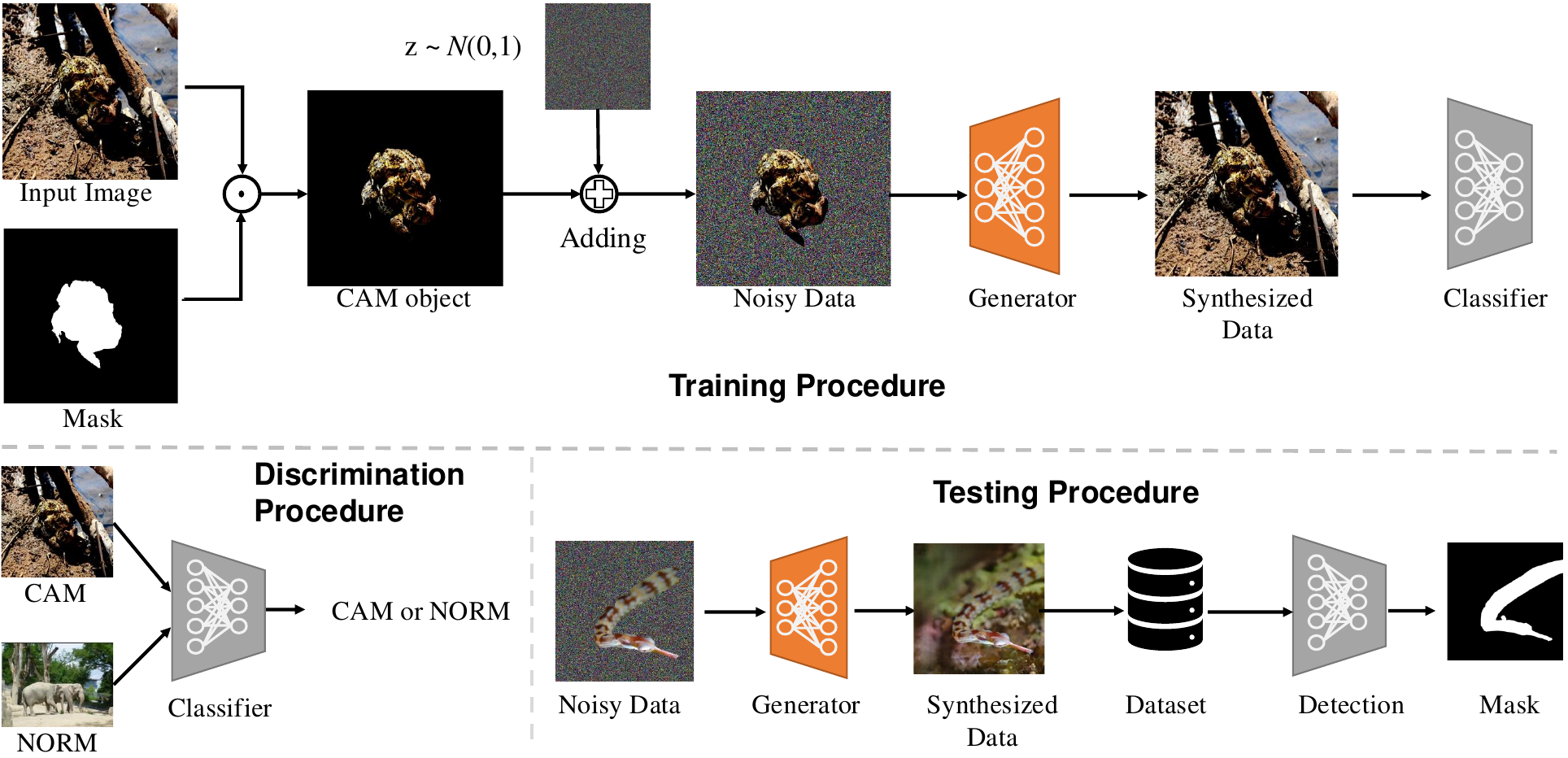}
\caption{\textbf{The architecture of our method.} 
% The upper part is our camouflage environment generator, while the left-down part is the training process of the Camouflage Classifier, and the right-down part is the dataset synthesis process. The camouflage environment generator takes an input foreground object and synthesizes the camouflage environment to help the foreground object hide in its background, which is camouflage. Please note that the "CAM" is short for camouflage, and "NORM" is short for normal.
The upper part shows our camouflage environment generator, which takes an input foreground object and synthesizes a camouflaged environment to help the object blend in with its background. The left-down part illustrates the training process of the Camouflage Classifier, and the right-down part shows the dataset synthesis process. Note that "CAM" denotes camouflage and "NORM" denotes normal images.
}  
%%\vspace{-15pt}
\label{fig:arch}
\end{figure*}
%-------------------------------------------------------------------------
% \section{Methodology}
% \label{sec:methods}
\section{Methodology}
\label{sec:methods}

In this section, we describe our approach to learning camouflaged object detection from synthetic data generated using generative adversarial networks (GANs). Given a sample from a segmentation dataset, we denote the input image as \(x\) (\(x \in \mathbb{R}^{H \times W \times C}\)) and the corresponding segmentation mask as \(y\) (\(y \in \mathbb{R}^{H \times W \times 1}\)). We sample a random noise vector \(z\) (\(z \in \mathbb{R}^{H \times W \times C}\)) from a standard normal distribution, where \(z\) has the same shape as \(x\).

The overview of our proposed approach, Synthetic Camouflage Detection (SCODE), is illustrated in Fig.~\ref{fig:arch}. We propose a camouflage environment generator to synthesize the expanded data \(\hat{x}\) from \(x\) and \(y\). We first introduce the notation of variables and then discuss each component in detail.

\subsection{Camouflage Distribution Loss}
\label{subsec:cdc}

Directly synthesizing a camouflage image is a challenging task because it requires supervision for the generator to learn the camouflage distribution. To address this issue, we propose a novel approach that leverages a classifier to provide supervision to the generation process.

\subsubsection{Understanding Camouflage Data}

Although humans can easily recognize a camouflage image, distinguishing between a camouflaged and a non-camouflaged image is challenging. Camouflage images are characterized by foreground objects that blend into their surrounding environment. This makes it difficult to mathematically express the relationship between the foreground object and the background environment.

Motivated by this consideration, we propose to train a classifier to tell the generator the distance between the synthesized results and real camouflage data. This classifier provides supervision to the generation process, enabling it to learn the camouflage distribution.

\subsubsection{Camouflage Classifier}

We train a classifier, denoted as \(C_{cam}\), to classify whether an image \(\hat{x}\) (synthetic or real) is a camouflage image or not. This process is defined as a binary classification problem. To train \(C_{cam}\), we use both natural and camouflage data. The architecture of \(C_{cam}\) is modified from the multi-scale discriminator in \cite{wang2018pix2pixHD}.

To learn the camouflage distribution, we leverage the softmax layer \cite{bridle1990probabilistic} and binary cross-entropy loss as the objective function. During the training process, \(C_{cam}\) learns to distinguish between natural and camouflage images, and in doing so, it learns the characteristics of camouflage images. We use the trained \(C_{cam}\) to provide supervision to the generation process. Specifically, we use the output of \(C_{cam}\) as the camouflage distribution loss \(\mathcal{L}_{cam}\). The loss function is defined as follows:
\begin{equation}
\mathcal{L}_{cam} = C_{cam}(\hat{x}),
\label{eq:cam_loss}
\end{equation}
where \(\hat{x}\) is the synthesized result of the generation process. By using the output of \(C_{cam}\) as the loss function, we ensure that the synthesized images are aligned with the camouflage distribution learned by \(C_{cam}\).

\subsection{Camouflage Environment Generation}
\label{subsec:ceg}

\subsubsection{Problem Definition}

Generating realistic camouflage data is a challenging task as traditional synthetic data methods struggle to learn the complex pixel distribution of objects with similar patterns and unclear boundaries against the background. Furthermore, due to the rarity of camouflage data, each object in the dataset is unique, making it difficult for generators to synthesize a suitable foreground object. Camouflage images require a precise match between the object and its environment; even a slight change in the background could disrupt the object's camouflage. Since camouflage objects do not alter their appearance in different environments, we disentangle camouflage images into two: foreground objects and background environments. By doing so, we redefine the problem as a camouflage background inpainting, which is closely related to the object's camouflage effect.

\subsubsection{Noise-Adding Module}

We describe our Noise-Adding Module for generating diverse camouflage background environments. We begin by sampling an image \(x\) and its segmentation mask \(y\) from the dataset. We then extract the foreground objects \(x_{fg}\) and background environment \(x_{bg}\) using the matrix dot product. Note that the mask \(y\) is normalized in the \([0,1]\) interval.

To increase the diversity of the background environments, we add a noise vector \(z\) to the background using a noise-adding process. Here, \(z\) is a sample from a standard normal distribution with the same shape as \(x\). The adding process can be formulated as follows:
\begin{equation}
x_{in} = x \odot y + z \odot (1-y),\quad z \in \mathcal{N}(0,1),
\label{eq:diffu}
\end{equation}
where \(x_{in}\) is the adding result and \(z\in\mathbb{R}^{H\times W\times C}\). Here, \(\odot\) denotes the dot product between two matrices.

\subsubsection{Camouflage Environment Generator}

We feed \(x_{in}\) into our camouflage environment generator \(G\) to synthesize camouflage images. To achieve high-resolution results, we adopt the Pix2PixHD~\cite{wang2018pix2pixHD} backbone. The generator outputs \(\hat{x}\), which is used to fool the discriminator \(D\).

Our multi-scale discriminator compares features extracted from different layers of the discriminator and is trained using adversarial training and feature-matching losses. The GAN loss encourages \(D\) to distinguish between real and generated images, while the feature matching loss minimizes the difference between features extracted from real and generated images. Specifically, the losses are defined as follows:
\begin{equation}\scriptsize
\begin{aligned}
\mathcal{L}_{\text{GAN}} = & \min_{G} \max_{D}  \mathbb{E}_{(x,x_{fg})}[\log D(x_{fg},x)] + \mathbb{E}_{x_{fg}}[\log (1- D(x_{fg},\hat{x}))],
\label{eqn::gan_problem}
\end{aligned}
\end{equation}
\begin{equation}\scriptsize
\begin{aligned}
\mathcal{L}_{\text{FM}}= \mathbb{E}_{(x,x_{fg})} \sum_{i=1}^T \frac{1}{N_i}[||D^{(i)}(x_{fg},x) -D^{(i)}(x_{fg},\hat{x})||_1],
\end{aligned}
\end{equation}
where \(D^{(i)}(*)\) is the feature of \(i\)-th layer when the input to \(D\) is \(*\). \(T\) and \(N_i\) are the total number of layers of \(D\) and the number of pixels in each layer, respectively.

The adversarial training loss \(\mathcal{L}_{G}\) of our camouflage environment generator is given by:
\begin{equation} 
\scriptsize
\mathcal{L}_{G} = \min_{G}\bigg(\Big(\max_{D} \sum \mathcal{L}_{\text{GAN}}\Big) + \lambda \sum\mathcal{L}_{\text{FM}}\bigg).
\label{eqn:multigan}
\end{equation}

To stabilize training, we use a VGG Loss~\cite{johnson2016perceptual} that calculates the perceptual distance between the feature maps of real and generated images. The VGG Loss is:
\begin{equation}\scriptsize
% \footnotesize
\mathcal{L}_{VGG} (\hat{x}, x) =  \frac1{C_jH_jW_j}\|\phi_j(\hat{x}) - \phi_j(x)\|_2^2,
\label{eqn:vgg}
\end{equation}
where \(\phi\) is a VGG network pretrained on ImageNet~\cite{Imagenet}. \(\phi_j(x)\) is the feature map of the \(j\)-th layer of the network \(\phi\) when processing the image \(x\).

The overall loss function of our camouflage environment generator is a weighted combination of the camouflage loss \(\mathcal{L}_{cam}\), GAN loss \(\mathcal{L}_{G}\), and VGG Loss \(\mathcal{L}_{VGG}\):
\begin{equation}
    \mathcal{L} = \lambda_{cam} \mathcal{L}_{cam} + \lambda_{G}  \mathcal{L}_{G} + \lambda_{VGG}  \mathcal{L}_{VGG},  %\lambda_{gram} * \ell_{gram}
    \label{eq:overall_loss}
\end{equation}
where the \(\lambda_*\) is the hyperparameter of \(\mathcal{L}_*\).

\subsection{Implementation Details}
\label{subsec:implementdetails}

In this subsection, we provide comprehensive details about the implementation of our proposed method. Our generator was trained using the Adam optimizer~\cite{kingma2015adam} with a learning rate set to 2e-4. The training process extended over 400 epochs, during which the learning rate remained constant for the initial 100 epochs and was then linearly decayed to zero. To accelerate training, we harnessed the computational power of four NVIDIA Tesla V100 GPUs, each equipped with 16GB of CUDA memory. A batch size of 16 was selected for training, and we resized the width and height of training images to 512 pixels.

For the detection models, we pursued a plug-and-play training approach on a single NVIDIA GeForce RTX 2080TI GPU, offering 11GB of CUDA memory. This approach was applied consistently in both the baseline models and the comparison experiments, where the baselines were either unaugmented or augmented by our SCODE model.

\begin{figure*}[h]
% %%\vspace{-22pt}
\begin{center}
\includegraphics[width=1\linewidth]{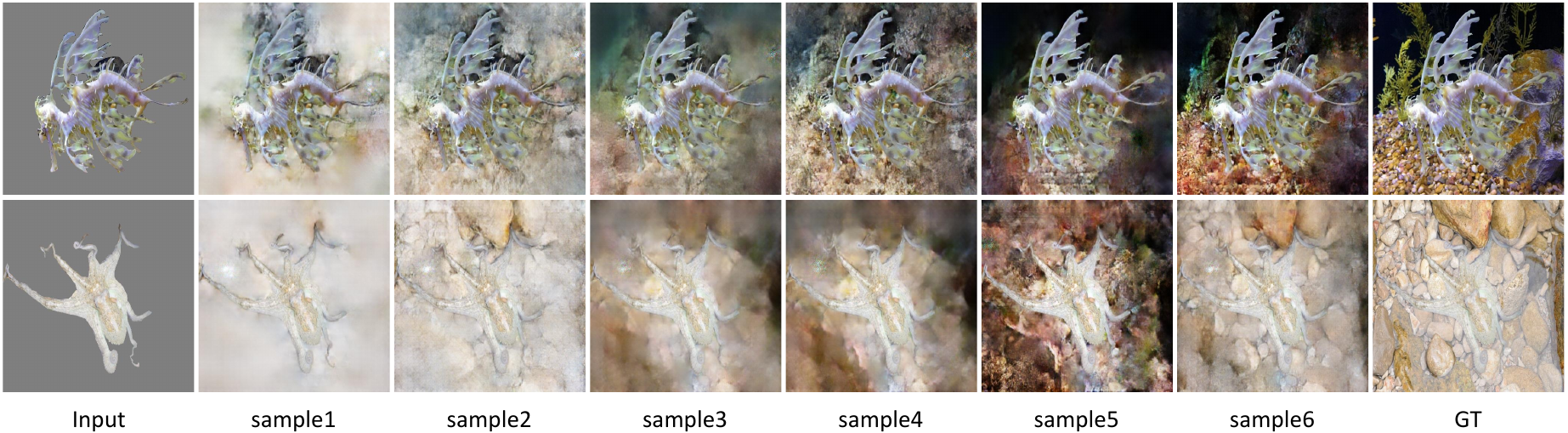}
\includegraphics[width=1\linewidth]{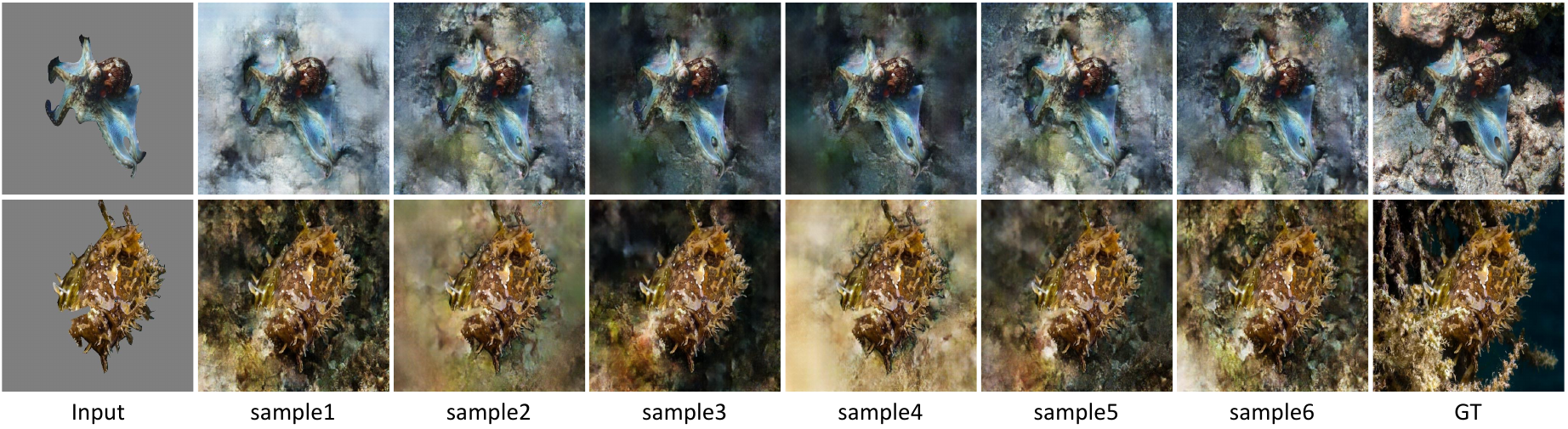}%%\vspace{-10pt}
\end{center}
  \caption{
 The Results of \textbf{Generative Diversity Experiment.} SCODE exhibits diverse results on both trained and untrained samples. \textbf{Because data augmentation should be applied on all images in the training set of detection models}, we visualize diversity results on the \textbf{training set} in the first two rows, and the \textbf{testing set} in the last two rows.
  }  %%\vspace{-20pt}
  \label{fig:diversity}
% %%\vspace{-10pt}
\end{figure*}

% \begin{figure*}[ht]
%   \centering
%   \includegraphics[width=1\textwidth]{fig/testvis.pdf}
%   %%\vspace{-5pt}
%   \caption{
%   Background Variation on Test Set. Backgrounds generated by our method exhibit variations compared to ground truth.
%   }  %%\vspace{-10pt}
%   \label{fig:testvis}
% \end{figure*}

\begin{figure*}[ht]
    \centering
    \includegraphics[width=1\textwidth]{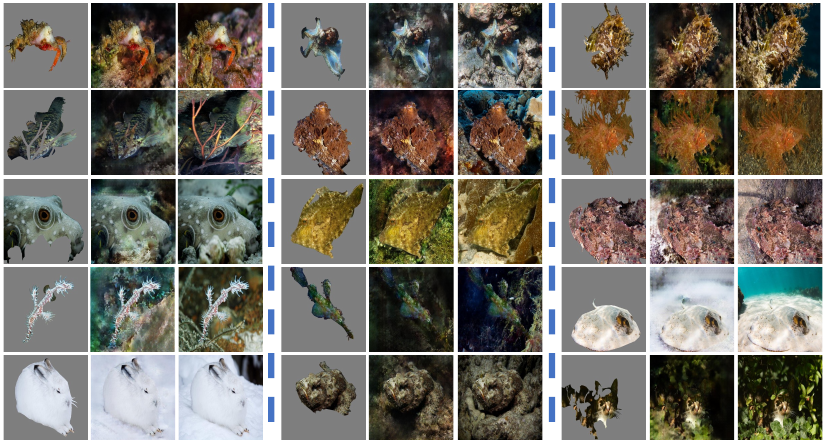}
        \vspace{0.5em} % Add some vertical spacing
    \begin{tabular}{@{}*{9}{>{\centering\arraybackslash}p{.09\textwidth}}@{}}
        Input & Out & GT & Input & Out & GT & Input & Out & GT \\
    \end{tabular}
    % Main caption for the entire figure
    \caption{
    \textbf{Background Variation on Test Set.} Backgrounds generated by our method exhibit variations compared to ground truth.
    }

    \label{fig:testvis}
\end{figure*}

\section{Experiments}
\label{sec:exp}
% In this section, we evaluate the effectiveness of our approach SCODE on camouflaged object detection in two aspects: detection and generation. We introduce the details of our baselines, metrics, and datasets in Sec.\ref{subsec:expsetting}. Then we introduce our SOTA quantitative detection results in Sec. \ref{subsec:quanres}. Qualitative results are conducted in Sec.\ref{subsec:qualres}, and its ablation study is in Sec.\ref{subsec:abl}. 
% Moreover, we visualize the mask consistency in \ref{subsec:mask}. We input out-of-distribution Non-camouflage objects from an unseen dataset to our model without training, and the out-of-distribution experiment is in Sec.\ref{subsec:ood}.

% In this section, we evaluate the effectiveness of our approach, SCODE, for camouflaged object detection in two aspects: detection and generation. 
In this section, we comprehensively evaluate the effectiveness of our proposed approach, SCODE, for camouflaged object detection. Our evaluation covers two key aspects: object detection performance and image generation quality.
% We describe the details of our baselines, metrics, and datasets in Sec.\ref{subsec:expsetting}. Then, we present our SOTA quantitative detection experiments results in Sec. \ref{subsec:quanres}. Our qualitative results and ablation study are presented in  Sec.\ref{subsec:qualres}, and the consistency of masks is examined in  Sec.~\ref{subsec:mask}. The ablation study is conducted in Sec.\ref{subsec:abl}.
% We also conduct an out-of-distribution experiment by inputting non-camouflaged objects from an unseen dataset to our model without training, which is presented in Sec.\ref{subsec:ood}.

\begin{figure*}[h]
    \centering
    \begin{subfigure}[t]{.99\columnwidth}
        \centering
        \includegraphics[width=\textwidth]{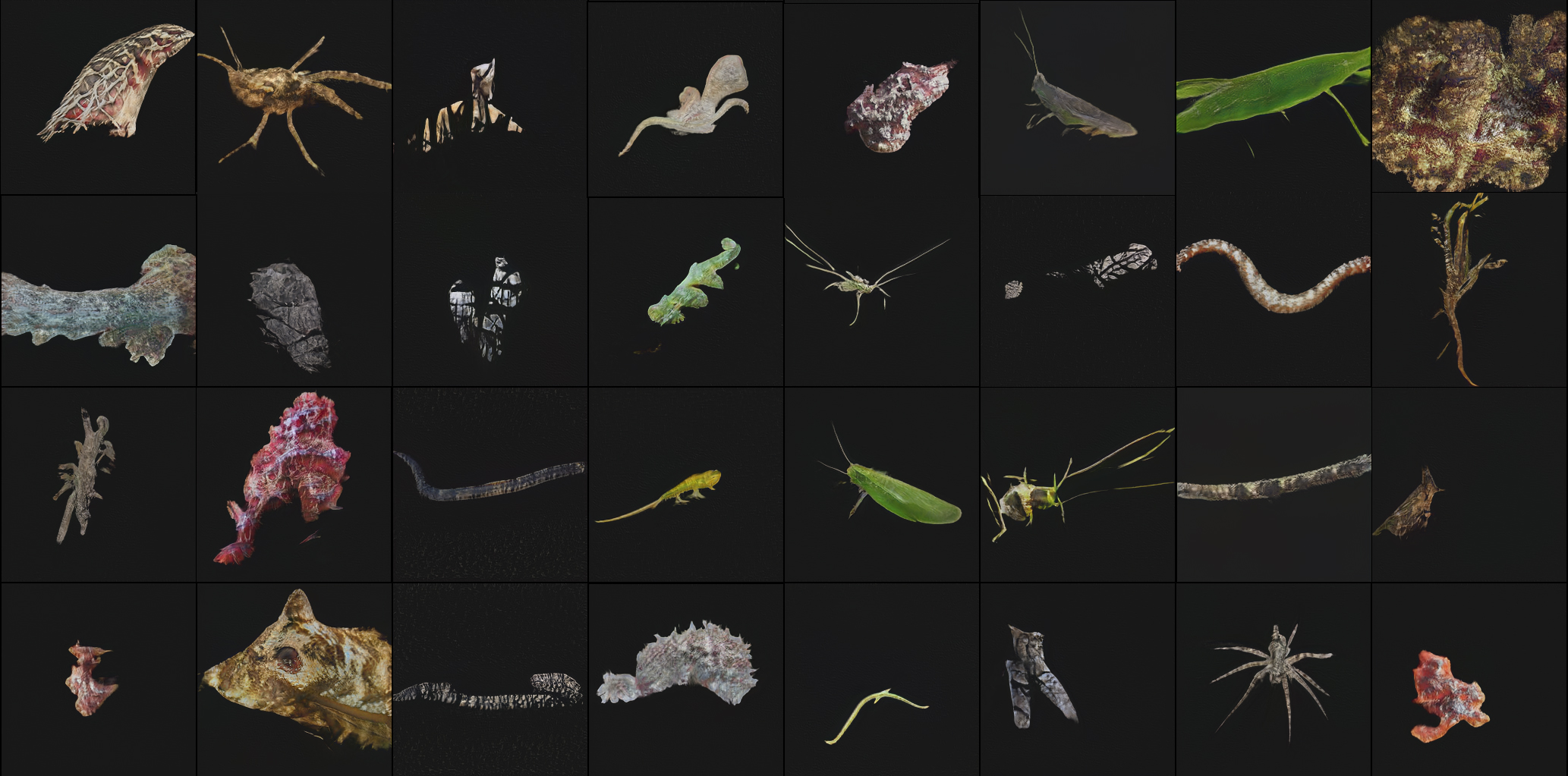}
        \caption*{Diverse Foreground Object Synthesis fed into SCODE} % No index for caption
    \end{subfigure}
    \hfill
    \begin{subfigure}[t]{.99\columnwidth}
        \centering
        \includegraphics[width=\textwidth]{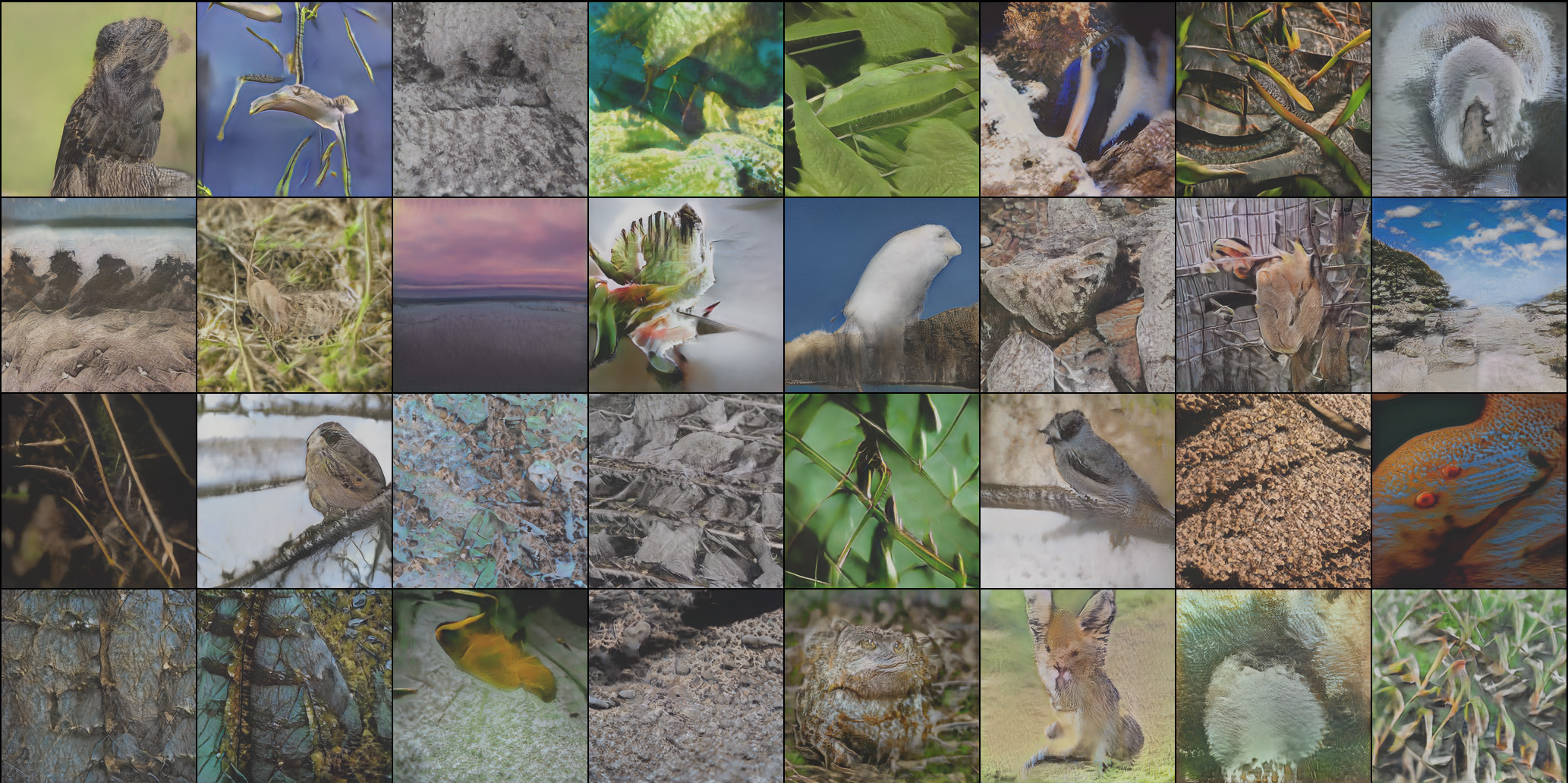}
        \caption*{StyleGAN can't synthesize entire camouflage image} % No index for caption
    \end{subfigure}
    \caption{Experiments on camouflaged foreground object synthesis}
    \label{fig:fore_gen}
\end{figure*}

% %-------------------------------------------------------------------------
\subsection{Experimental Settings}
\label{subsec:expsetting}
% Following the common camouflaged object detection protocol, we use three classic camouflaged object detection datasets, the CAMO~\cite{le2019anabranch},
% CHAMELEON~\cite{2018Animal}, and COD10K~\cite{fan2020camouflaged}. 
% We train our SCODE method on these three datasets and use synthetic data to extend the training dataset, which is then used to train the camouflaged object detection method. Regarding camouflaged object detection, we follow the same experiment settings as ~\cite{fan2021concealed}.

% We use three camouflaged object detection datasets for our experiments: CAMO~\cite{le2019anabranch}, CHAMELEON~\cite{2018Animal}, and COD10K~\cite{fan2020camouflaged}. We train our SCODE method on these three datasets and use synthetic data to augment the training dataset, which is then used to train our camouflaged object detection method. Following the experimental settings as ~\cite{fan2021concealed} for camouflaged object detection, we combine the three training parts as training datasets and test the model separately on each testing dataset. We generate an augmented image for each image in the training dataset by the SCODE method unless otherwise specified. The details about the dataset, dataset splitting, and metrics can be found in the supplementary materials.

We conducted our experiments on three distinct camouflaged object detection datasets: CAMO~\cite{le2019anabranch}, CHAMELEON~\cite{2018Animal}, and COD10K~\cite{fan2020camouflaged}. Our approach involves training the SCODE model on these datasets and utilizing synthetic data to augment the training data. This augmented dataset is then employed to train our camouflaged object detection model. Following the experimental setup outlined in~\cite{fan2021concealed}, we combined the training subsets from the three datasets to form the training dataset, while the performance was evaluated individually on the respective test subsets. SCODE-generated images were added to the training dataset through data augmentation, unless otherwise specified. Additional details regarding dataset specifics, partitioning, and performance metrics can be found in the supplementary materials.

% %%\vspace*{-10pt}
%-------------------------------------------------------------------------
\subsubsection{Baselines}
% We select 14 deep learning models ~\cite{lin2017feature, he2017mask, zhao2017pyramid, zou2018DLMIA, liu2018picanet, huang2019mask, zhao2019pyramid, wu2019cascaded, chen2019hybrid, le2019anabranch, zhao2019EGNet, fan2020pranet, fan2020camouflaged, fan2021concealed} as our baselines for camouflaged object detection. 
% And we select three recent baselines ZoomNet~\cite{ZoomNet-CVPR2022}, SINet-V2~\cite{fan2021concealed}, FEDER~\cite{He2023Camouflaged} in the plug-and-play experiment.
% Since the goal of our SCODE method is to improve the performance of camouflaged object detection through synthetic data augmentation, although no publicly available work focuses on synthesizing camouflage data, we choose detection performance as our evaluation criterion to assess the performance of our SCODE method.

We selected 14 deep learning models~\cite{lin2017feature, he2017mask, zhao2017pyramid, zou2018DLMIA, liu2018picanet, huang2019mask, zhao2019pyramid, wu2019cascaded, chen2019hybrid, le2019anabranch, zhao2019EGNet, fan2020pranet, fan2020camouflaged, fan2021concealed} as our baselines for camouflaged object detection. Additionally, we included three recent baselines, namely ZoomNet~\cite{ZoomNet-CVPR2022}, SINet-V2~\cite{fan2021concealed}, and FEDER~\cite{He2023Camouflaged}, in the plug-and-play experiment. As our SCODE method aims to enhance camouflaged detection by synthetic data augmentation, we focused on detection performance as our evaluation criterion, in alignment with the context of our approach.

% %-------------------------------------------------------------------------
% \subsubsection{Experimental Settings}
% \label{subsec:expsetting}
% % Following the common camouflaged object detection protocol, we use three classic camouflaged object detection datasets, the CAMO~\cite{le2019anabranch},
% % CHAMELEON~\cite{2018Animal}, and COD10K~\cite{fan2020camouflaged}. 
% % We train our SCODE method on these three datasets and use synthetic data to extend the training dataset, which is then used to train the camouflaged object detection method. Regarding camouflaged object detection, we follow the same experiment settings as ~\cite{fan2021concealed}.

% We use three camouflaged object detection datasets for our experiments: CAMO~\cite{le2019anabranch}, CHAMELEON~\cite{2018Animal}, and COD10K~\cite{fan2020camouflaged}. We train our SCODE method on these three datasets and use synthetic data to augment the training dataset, which is then used to train our camouflaged object detection method. Following the experimental settings as ~\cite{fan2021concealed} for camouflaged object detection, we combine the three training parts as training datasets and test the model separately on each testing dataset. We generate an augmented image for each image in the training dataset by the SCODE method unless otherwise specified.

%-------------------------------------------------------------------------
\begin{table*}[t!]%\scriptsize
  \centering
  \caption{\textbf{Quantitative results on three different datasets.} 
    The best scores are highlighted in \textbf{bold}. Our model outperforms other 14 baselines.
    Note that the ANet-SRM model (only trained on CAMO) does not have 
    a publicly available code, 
    thus other results are not available. 
    $\uparrow$ and $\downarrow$ indicate higher scores better and lower score better, respectively.
    %For DDCN, we use the pre-trained model due to the released code has error.
    % $E_\phi$ denotes mean E-measure~\cite{Fan2018Enhanced}. 
    %\cmm{Baseline models are trained using the training settings in our 
    %conference version~\cite{fan2020camouflaged}.} 
  }\label{tab:ModelScore} %%\vspace{-10pt}
 
  % %%\vspace{-8pt}

  \begin{tabular}{l|cccc||cccc||cccc} \hline \toprule 
    & \multicolumn{4}{c||}{{CHAMELEON~\cite{2018Animal}}} 
    & \multicolumn{4}{c||}{{CAMO-Test~\cite{le2019anabranch}}}
    & \multicolumn{4}{c  }{{COD10K-Test~\cite{fan2020camouflaged}}} 
    \\ \cline{2-13}
    Baseline Models~~~ 
    & $S_\alpha\uparrow$ &$E_\phi\uparrow$ &$F_\beta^w\uparrow$ &$M\downarrow$
    & $S_\alpha\uparrow$ &$E_\phi\uparrow$ &$F_\beta^w\uparrow$ &$M\downarrow$
    %&$S_\alpha\uparrow$ &$E_\phi\uparrow$ &$F_\beta^w\uparrow$ &$M\downarrow$
    & $S_\alpha\uparrow$ &$E_\phi\uparrow$ &$F_\beta^w\uparrow$ &$M\downarrow$ 
    \\ \hline
    \lach
    FPN~\cite{lin2017feature}
    &0.794&0.783&0.590&0.075&0.684&0.677&0.483&0.131&0.697&0.691&0.411&0.075\\
    \lach
    MaskRCNN~\cite{he2017mask}
    &0.643&0.778&0.518&0.099&0.574&0.715&0.430&0.151&0.613&0.748&0.402&0.080\\
    \lach
    PSPNet~\cite{zhao2017pyramid}
    &0.773&0.758&0.555&0.085&0.663&0.659&0.455&0.139&0.678&0.680&0.377&0.080\\
    \lach
    UNet++~\cite{zou2018DLMIA}
    &0.695&0.762&0.501&0.094&0.599&0.653&0.392&0.149&0.623&0.672&0.350&0.086\\
    \lach
    PiCANet~\cite{liu2018picanet}
    &0.769&0.749&0.536&0.085&0.609&0.584&0.356&0.156&0.649&0.643&0.322&0.090\\
    \lach
    MSRCNN~\cite{huang2019mask}
    &0.637&0.686&0.443&0.091&0.617&0.669&0.454&0.133&0.641&0.706&0.419&0.073\\
    \lach
    PFANet~\cite{zhao2019pyramid}
    &0.679&0.648&0.378&0.144&0.659&0.622&0.391&0.172&0.636&0.618&0.286&0.128\\
    \lach
    CPD~\cite{wu2019cascaded}
    &0.853&0.866&0.706&0.052&0.726&0.729&0.550&0.115&0.747&0.770&0.508&0.059\\ 
    \lach
    HTC~\cite{chen2019hybrid}
    &0.517&0.489&0.204&0.129&0.476&0.442&0.174&0.172&0.548&0.520&0.221&0.088\\
    \lach
    ANet-SRM~\cite{le2019anabranch}
    & - & - & - & - &0.682&0.685&0.484&0.126& - & - & - & -\\
    \lach
    EGNet~\cite{zhao2019EGNet}
    &0.848&0.870&0.702&0.050&0.732&0.768&0.583&0.104&0.737&0.779&0.509&0.056\\
    \lach
    PraNet~\cite{fan2020pranet}     
    &0.860&0.907&0.763&0.044&0.769&0.824&0.663&0.094&0.789&0.861&0.629&0.045\\
    \lach
    SINet~\cite{fan2020camouflaged} & {0.869} & {0.891} & {0.740} & {0.044} & {0.751} & {0.771} & {0.606} & {0.100} & {0.771} & {0.806} & {0.551} & {0.051} \\
    \lach
    SINet-V2~\cite{fan2021concealed}
    & {0.888} & {0.942} & {0.816} & {0.030}
    & {0.811} & {0.873} & {0.730} & {0.072}
    & {0.815} & {0.887} & {0.680} & {0.037} \\ 
    \hline
    \lach
    \textbf{\emph{SCODE}~(Ours)}
    & \textbf{0.892} & \textbf{0.959} & \textbf{0.822} & \textbf{0.029}
    & \textbf{0.813} & \textbf{0.882} & \textbf{0.736} & \textbf{0.071}
    & \textbf{0.815} & \textbf{0.902} & \textbf{0.684} & \textbf{0.036} 
    \\ \bottomrule
  \end{tabular}
\end{table*}

\begin{table*}[htbp]
%%\vspace{-10pt}
\centering
  \caption{\textbf{Plug-and-Play Quantitative results on three different datasets.} 
    The best scores are highlighted in \textbf{bold}. Our model outperforms when plugged into three baselines.
    % $\uparrow$ indicates higher score is better, and $\downarrow$ indicates the lower score is better.
    %For DDCN, we use the pre-trained model due to the released code has error.
    % $E_\phi$ denotes mean E-measure~\cite{Fan2018Enhanced}. 
    %\cmm{Baseline models are trained using the training settings in our 
    %conference version~\cite{fan2020camouflaged}.} 
  }\label{tab:AddScore} %%\vspace{-10pt}
\resizebox{1\linewidth}{!}{
\resizebox{1\linewidth}{!}{
\begin{tabular}{l|cccc||cccc||cccc} \hline \toprule 
    & \multicolumn{4}{c||}{{CHAMELEON}} 
    & \multicolumn{4}{c||}{{CAMO-Test}}
    & \multicolumn{4}{c  }{{COD10K-Test}} 
    \\ \cline{2-13}
    Baseline Models~~~ 
    & $S_\alpha\uparrow$ &$E_\phi\uparrow$ &$F_\beta^w\uparrow$ &$M\downarrow$
    & $S_\alpha\uparrow$ &$E_\phi\uparrow$ &$F_\beta^w\uparrow$ &$M\downarrow$
    %&$S_\alpha\uparrow$ &$E_\phi\uparrow$ &$F_\beta^w\uparrow$ &$M\downarrow$
    & $S_\alpha\uparrow$ &$E_\phi\uparrow$ &$F_\beta^w\uparrow$ &$M\downarrow$ 
    \\ \hline
    ZoomNet~\cite{ZoomNet-CVPR2022}
    & {0.763} & {0.854} & {0.662} & {0.083}
    & {0.763} & {0.801} & {0.662} & {0.083}
    & {0.804} & {0.843} & {0.672} & {0.034} \\ 
    % \hline
    \lach
    \textbf{Ours+ZoomNet}
    & \textbf{0.833} & \textbf{0.876} & \textbf{0.749} & \textbf{0.037}
    & \textbf{0.780} & \textbf{0.866} & \textbf{0.683} & \textbf{0.081}
    & \textbf{0.812} & \textbf{0.889} & \textbf{0.679} & \textbf{0.033} \\ 
    \hline
    \lach
    SINet-V2~\cite{fan2021concealed}
    & {0.888} & {0.942} & {0.816} & {0.030}
    & {0.811} & {0.873} & {0.730} & {0.072}
    & {0.815} & {0.887} & {0.680} & {0.037} \\ 
    % \hline
    \lach
    \textbf{Ours+SINet-V2}
    & \textbf{0.892} & \textbf{0.959} & \textbf{0.822} & \textbf{0.029}
    & \textbf{0.813} & \textbf{0.882} & \textbf{0.736} & \textbf{0.071}
    & \textbf{0.815} & \textbf{0.902} & \textbf{0.684} & \textbf{0.036} \\ 
    \hline
    \lach
    FEDER~\cite{He2023Camouflaged}
    & {0.599} & {0.757} & {0.363} & {0.130}
    & {0.555} & {0.717} & {0.466} & {0.168}
    & {0.566} & {0.670} & {0.246} & {0.109} \\ 
    % \hline
    \lach
    \textbf{Ours+FEDER}
    & \textbf{0.680} & \textbf{0.821} & \textbf{0.477} & \textbf{0.097}
    & \textbf{0.558} & \textbf{0.725} & \textbf{0.505} & \textbf{0.152}
    & \textbf{0.597} & \textbf{0.712} & \textbf{0.308} & \textbf{0.092} 
    \\ \bottomrule
  \end{tabular}
}}%%\vspace{-10pt}
\end{table*}

% \textbf{Comparison with Data Augmentation Methods.}
% Thanks. We compare our SCODE with non-deep learning-based data augmentation techniques in this table.
\begin{table*}[ht]
\centering
  \caption{\textbf{Comparison with Data Augmentation Methods.} 
    We compare our SCODE with non-deep learning-based data augmentation techniques in this table.
    $\uparrow$ indicates higher score is better, and $\downarrow$ indicates the lower score is better.
  }\label{tab:Comparison_aug} 
% %%\vspace{-10pt}
\resizebox{1.0\linewidth}{!}{
\begin{tabular}{l|cccc||cccc||cccc} \hline \toprule 
    & \multicolumn{4}{c||}{{CHAMELEON}} 
    & \multicolumn{4}{c||}{{CAMO-Test}}
    & \multicolumn{4}{c  }{{COD10K-Test}} 
    \\ \cline{2-13}
    Data Augmentation~~~ 
    & $S_\alpha\uparrow$ &$E_\phi\uparrow$ &$F_\beta^w\uparrow$ &$M\downarrow$
    & $S_\alpha\uparrow$ &$E_\phi\uparrow$ &$F_\beta^w\uparrow$ &$M\downarrow$
    %&$S_\alpha\uparrow$ &$E_\phi\uparrow$ &$F_\beta^w\uparrow$ &$M\downarrow$
    & $S_\alpha\uparrow$ &$E_\phi\uparrow$ &$F_\beta^w\uparrow$ &$M\downarrow$ 
    \\ \hline
    SINet-V2 + Random Crop
    & {0.873} & {0.920} & {0.796} & {0.036}
    & {0.789} & {0.842} & {0.697} & {0.081}
    & {0.798} & {0.858} & {0.651} & {0.040} \\ 
    % \hline
    \lach
    SINet-V2 +Random Rotation
    & {0.879} & {0.924} & {0.801} & {0.036}
    & {0.804} & {0.858} & {0.716} & {0.078}
    & {0.798} & {0.850} & {0.649} & {0.041} \\ 
    % \hline
    \lach
    \textbf{SINet-V2 + SCODE (Ours)}
    & \textbf{0.892} & \textbf{0.959} & \textbf{0.822} & \textbf{0.029}
    & \textbf{0.813} & \textbf{0.882} & \textbf{0.736} & \textbf{0.071}
    & \textbf{0.815} & \textbf{0.902} & \textbf{0.684} & \textbf{0.036}
    \\ \bottomrule
  \end{tabular}
}%%\vspace{-10pt}
\end{table*}

% \begin{figure}[t]
%   \centering
%   \includegraphics[width=.50\textwidth]{picture/qual_compare.pdf}
%   % %%\vspace*{-15pt}
%   \caption{\textbf{Comparable Qualitative Experiments}.  We compare our model with two baseline models, CycleGAN~\cite{CycleGAN2017} and pix2pixHD~\cite{wang2018pix2pixHD}, on the task of synthesizing camouflage images. The results show that while the baseline models exhibit artifacts and inconsistencies when trained on camouflaged datasets, our model is able to synthesize realistic camouflage images with accurate background textures and patterns. Please note that here we use the training set of SCODE to synthesize similar backgrounds and better compare with ground truth.
%   }  
%   %%\vspace{-10pt}
%   \label{fig:qual_compare}
% \end{figure}

\begin{figure}[t]
    \centering
    \includegraphics[width=.50\textwidth]{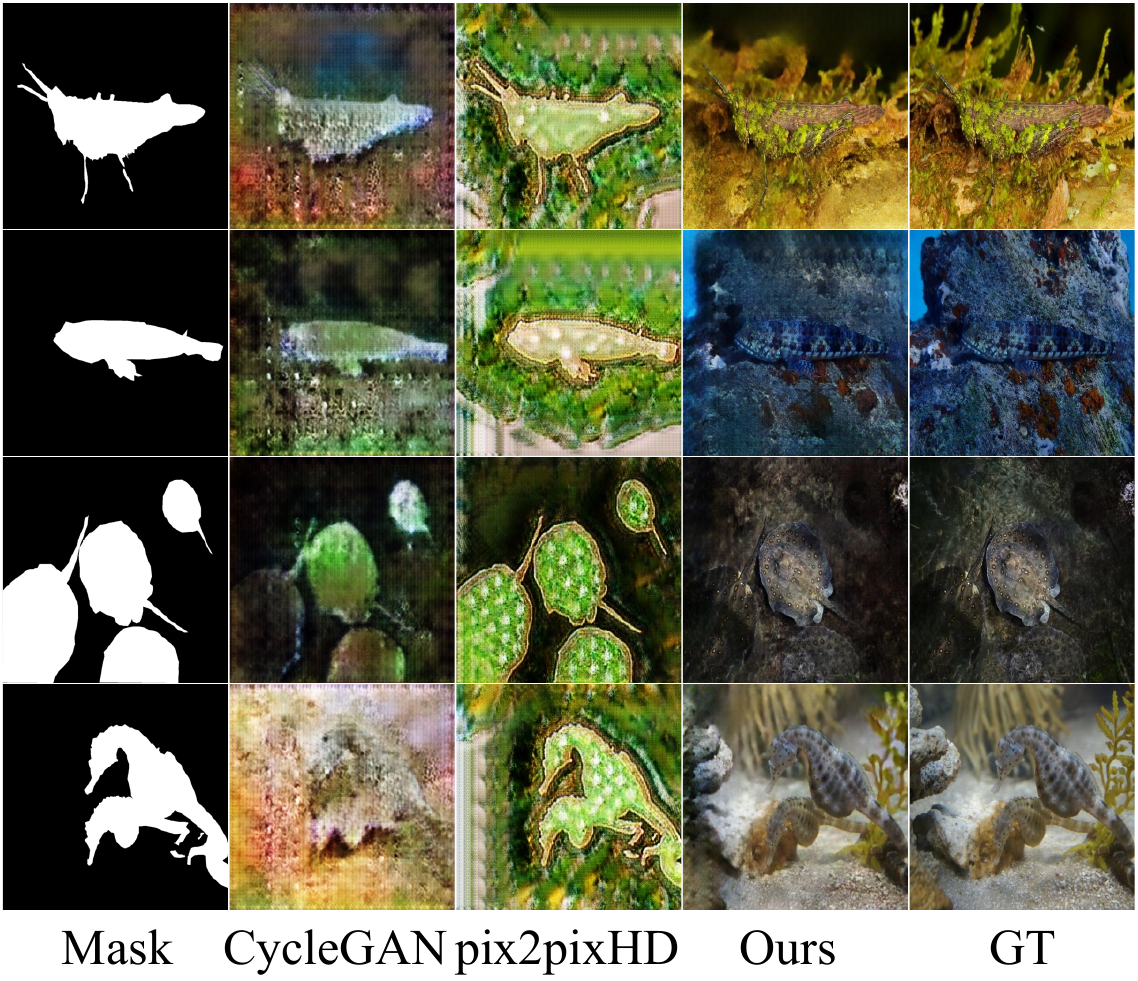}
        \vspace{0.5em} % Add some vertical spacing
    \begin{tabular}{@{}*{5}{>{\centering\arraybackslash}p{.08\textwidth}}@{}}
        Mask & CycleGAN & pix2pixHD & Ours & GT \\
    \end{tabular}
    % Main caption for the entire figure
    \caption{\textbf{Comparable Qualitative Experiments}. We compare our model with two baseline models, CycleGAN~\cite{CycleGAN2017} and pix2pixHD~\cite{wang2018pix2pixHD}, on the task of synthesizing camouflage images. The results show that while the baseline models exhibit artifacts and inconsistencies when trained on camouflaged datasets, our model is able to synthesize realistic camouflage images with accurate background textures and patterns. Please note that here we use the training set of SCODE to synthesize similar backgrounds and better compare with ground truth.}

    \label{fig:qual_compare}
\end{figure}

\subsection{Datasets}
We evaluate our method on three widely used camouflaged object detection datasets. Since these datasets do not have large amounts of data, our data generation methods are essential to expanding these datasets.

% \noindent\textbf{CAMO~\cite{le2019anabranch}}
\subsubsection{CAMO~\cite{le2019anabranch}}
% The camouflage Object (CAMO) dataset is a camouflaged object segmentation dataset on two categories of data, \ie, naturally camouflaged objects and artificially camouflaged objects. The CAMO contains 1250 camouflaged object images (1000 images in the training set and 250 images in the test set) and non-camouflaged objects (1000 images in the training set and 250 images in the test set). The two kinds of data are collected from real-world animals and humans. 
This dataset contains 1250 images of naturally and artificially camouflaged objects, with 1000 images in the training set and 250 images in the test set. It contains 1000 non-camouflaged object images in the training set and 250 non-camouflaged object images in the test set. The dataset was collected from real-world animals and humans.

% \noindent\textbf{CHAMELEON ~\cite{2018Animal}}
\subsubsection{CHAMELEON ~\cite{2018Animal}}
CHAMELEON is an unpublished datset collected from google search. It has only 76 images with manually annotated ground truth. This dataset consists of 76 images with manually annotated ground truth. Although it is an old dataset, it is widely used as a benchmark for camouflaged object detection. However, it can only be used in testing because the quantity of images is not enough to train deep-learning models.

% \noindent\textbf{COD10K~\cite{fan2020camouflaged}}
\subsubsection{COD10K~\cite{fan2020camouflaged}}
% COD10K is a camouflage dataset collected by Fan et al.  in 2020. Although it contains 10000 images (6000 for the training set and 4000 for the testing set), it has only 5,066 camouflaged images. The others are 3,000 background images and 1,934 non-camouflaged images. It has ten primary classes and 78 sub-classes. The data source is from photography websites.
This dataset, collected by Fan et al.  ~\cite{fan2020camouflaged} in 2020, contains 10000 images, of which 5066 images are camouflaged, 3000 images are backgrounds, and 1934 images are non-camouflaged. It has ten primary classes and 78 sub-classes. The data source is from photography websites.

\begin{table*}[h]

\centering
\caption{The Quantitative Results of \textbf{Ablation Study on CAM.} }
% \vspace{-10pt}
\resizebox{1\linewidth}{!}{

\begin{tabular}{l|cccc||cccc||cccc} \hline \toprule 
    & \multicolumn{4}{c||}{{CHAMELEON}} 
    & \multicolumn{4}{c||}{{CAMO-Test}}
    & \multicolumn{4}{c  }{{COD10K-Test}} 
    \\ \cline{2-13}
    Module Variant~~~ 
    & $S_\alpha\uparrow$ &$E_\phi\uparrow$ &$F_\beta^w\uparrow$ &$M\downarrow$
    & $S_\alpha\uparrow$ &$E_\phi\uparrow$ &$F_\beta^w\uparrow$ &$M\downarrow$
    %&$S_\alpha\uparrow$ &$E_\phi\uparrow$ &$F_\beta^w\uparrow$ &$M\downarrow$
    & $S_\alpha\uparrow$ &$E_\phi\uparrow$ &$F_\beta^w\uparrow$ &$M\downarrow$ 
    \\ \hline
    w/o CAM
    & {0.886} & {0.919} & {0.812} & {0.032}
    & {0.808} & {0.867} & {0.812} & {0.075}
    & {0.814} & {0.860} & {0.676} & {0.037} \\ 
    % \hline
    \lach
    \textbf{whole model}
    & \textbf{0.892} & \textbf{0.959} & \textbf{0.822} & \textbf{0.029}
    & \textbf{0.813} & \textbf{0.882} & \textbf{0.736} & \textbf{0.071}
    & \textbf{0.815} & \textbf{0.902} & \textbf{0.684} & \textbf{0.036}
    \\ \bottomrule
  \end{tabular}
  % \vspace{-22pt}
    \label{tab:ablaquan}
}
\end{table*}

\begin{table}[b]
\caption{The Number of Augmented Samples }
\centering
\resizebox{1\linewidth}{!}{
\begin{tabular}{l|ccc} \hline \toprule 
    Dataset Partition~~~ 
    & \multicolumn{2}{c}{\tabincell{c}{Training Set(CAMO + COD10k)}} & Test Set(Each Dataset)
    \\ \hline
    % Module Variant~~~ 
    % & $S_\alpha\uparrow$ &$E_\phi\uparrow$ &$F_\beta^w\uparrow$ &$M\downarrow$
    % \\ \hline
    SCODE Augmented~~~
    & Before & After  & - \\ 
    \hline
    \lach
    \textbf{\#Number of Data~~~}
    & \textbf{4040} & \textbf{8080}   & \textbf{unchange}  
    \\ \bottomrule
  \end{tabular}
  \label{tab:num}
}
\end{table}

%-------------------------------------------------------------------------
% \vspace*{-10pt}
\subsection{Metrics}
We use four commonly used metrics for evaluating camouflaged object detection methods.

% \noindent\textbf{S-measure($S_\alpha$)~\cite{fan2017structure}}
\subsubsection{S-measure($S_\alpha$)~\cite{fan2017structure}}
The Structure-measure (S-measure) is a segmentation evaluation metric. Instead of calculating pixel-wise errors, it evaluates the region-aware and object-aware structural similarity of non-binary foreground maps.

% \noindent\textbf{E-measure($E_\phi$)~\cite{Fan2018Enhanced}}
\subsubsection{E-measure($E_\phi$)~\cite{Fan2018Enhanced}}

The E-measure (Enhanced-alignment Measure) is proposed by Fan et al.  for binary foreground evaluation. Imitating human perception of sensitivity to both global
information and local details, it assesses the image-level and pixel-level accuracy of the camouflaged object detection results. It is a human visual perception evaluation metric.

% \noindent\textbf{weighted F-measure($F_\beta^w$)~\cite{margolin2014evaluate} }
\subsubsection{Weighted F-measure($F_\beta^w$)~\cite{margolin2014evaluate}}
The weighted F-measure is a segmentation metric for both non-binary maps and binary maps. It was proposed by Margolin et al.  in 2014. It is a weighted harmonic mean between recall and precision. Some recent works ~\cite{Fan2018Enhanced,fan2017structure,fan2020camouflaged,fan2021concealed} suggested the weighted F-measure is more reliable than the traditional $F_\beta$.

% \noindent\textbf{Mean Absolute Error($M$)~\cite{perazzi2012saliency}}
\subsubsection{Mean Absolute Error($M$)~\cite{perazzi2012saliency}}
The MAE (Mean Absolute Error) is a common metric for segmentation tasks to assess the presence and amount of error. It exhibits $\ell_1$ loss between ground truth and segmentation. However, it is not sensitive to the location of pixels because it is too average to determine where such errors occur.
% \vspace{-7pt}
%-------------------------------------------------------------------------
% \subsection{Dataset Split}
% For the training and testing of our SCODE data augmentation method, we have split the camouflaged object detection (COD) dataset. SCODE is our data augmentation technique that requires training and inference on the training set of detection methods to generate augmented samples. That means, the training set of the COD model is the dataset (including the training and testing set) of the SCODE. To ensure the separation of SCODE from the testing set to avoid data leaking in COD, we split the COD training set for the training and testing of SCODE. In Fig.~\ref{fig:qulares}, we show the augmented trained samples; Fig.~\ref{fig:qulr} showcases augmented testing samples. The results in Table 1 in the main paper demonstrate that even on the trained samples, the SCODE augmentation introduces varieties and differences compared to the original dataset. These variations contribute to the improvement of the COD method through effective data augmentation. By carefully isolating the COD testing sets, we ensure the efficacy of SCODE and provide evidence of its impact on enhancing COD models through many quantitative results.

\begin{figure*}[h]
    \centering
    \includegraphics[width=1\linewidth]{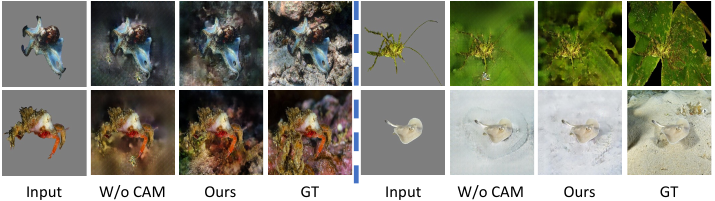}
    \vspace{0.5em} % Add vertical space
    % Custom table for captions
    \begin{tabular}{@{}*{8}{>{\centering\arraybackslash}p{.10\textwidth}}@{}}
        Input & W/o CAM & Ours & GT & Input & W/o CAM & Ours & GT \\
    \end{tabular}
    \includegraphics[width=1\textwidth]{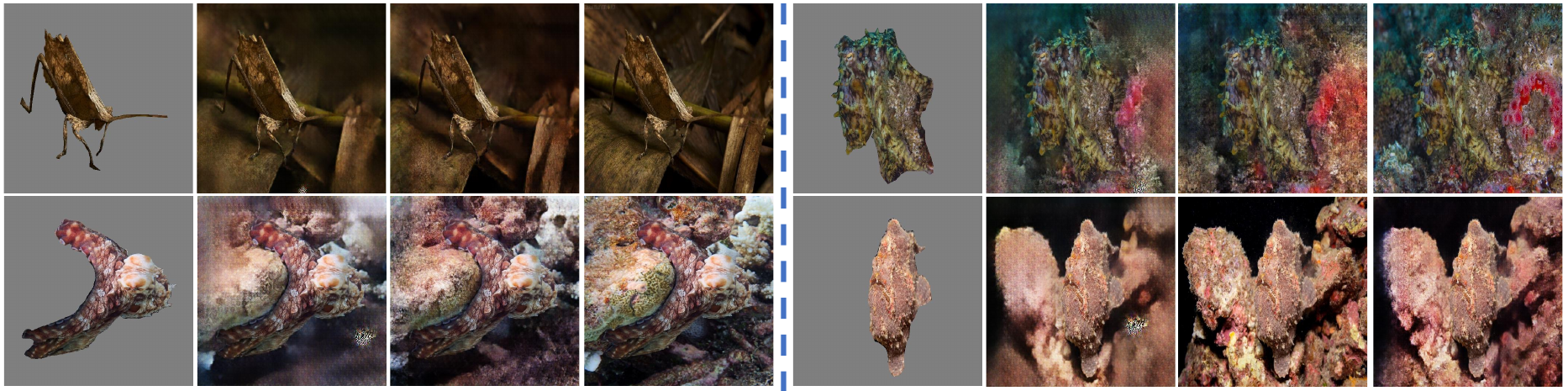}
    \vspace{0.5em} % Add vertical space
    % Custom table for captions
    \begin{tabular}{@{}*{8}{>{\centering\arraybackslash}p{.10\textwidth}}@{}}
        Input & W/o Adding & Ours & GT & Input & W/o Adding & Ours & GT \\
    \end{tabular}
    
    % Main caption for the entire figure
    \caption{
    \textbf{The Results of Ablation Study on Noise Adding and CAM.} Note that the "w/o Adding" and "w/o CAM" refer to results without the noise-adding process and camouflage classifier, respectively. The results show that without our noise-adding module, the model tends to produce blurred backgrounds; without CAM, the model struggles to synthesize details in the camouflaged background.
    }

    \label{fig:ablation}
\end{figure*}
\subsection{Dataset Split}
For the training and testing of our SCODE data augmentation method, we have split the camouflaged object detection (COD) dataset. SCODE is our data augmentation technique that requires training and inference on the training set of detection methods to generate augmented samples. That means the training set of the COD model is the dataset (including the training and testing set) of the SCODE. To ensure the separation of SCODE from the testing set to avoid data leaking in COD, we split the COD training set for the training and testing of SCODE. 

% In Fig.~\ref{fig:qulares}, we show the augmented trained samples; 
Fig.~\ref{fig:qulr} showcases augmented testing samples. The results in Table ~\ref{tab:ablaquan} in the main paper demonstrate that even on the trained samples, the SCODE augmentation introduces varieties and differences compared to the original dataset. These variations contribute to the improvement of the COD method through effective data augmentation. By carefully isolating the COD testing sets, we ensure the efficacy of SCODE and provide evidence of its impact on enhancing COD models through many quantitative results.

% \vspace{-10pt}
%-------------------------------------------------------------------------
% \subsection{Ablation Study}
% \label{subsec:abl}

% To evaluate the effectiveness of our method, in addition to the qualitative experiment in the main paper, we conducted an ablation study to see the quantitative results of CAM in SCODE. 
% As shown in Table.~\ref{tab:ablaquan}, we observed that without the camouflage distribution classifier, the performance drops a little bit compared to the whole model. Combined with the results in Figure 5 in the main paper, it can be seen that the performance drop is due to the difficulty of the model to synthesize backgrounds with details. 
\subsection{Quantitative Ablation Study}
\label{supp:subsec:abl}

In addition to the qualitative experiment presented in the main paper, we conducted an ablation study to assess the quantitative impact of the camouflage distribution classifier (CAM) in SCODE. The results are summarized in Table~\ref{tab:ablaquan}.

As shown in Table~\ref{tab:ablaquan}, we observed that when excluding the camouflage distribution classifier (CAM), the performance slightly decreased compared to the full SCODE model. This decline in performance suggests that the CAM module plays a crucial role in improving the model's ability to synthesize backgrounds with fine details. Combined with the qualitative results presented in Figure 5 in the main paper, it can be inferred that the performance drop is primarily due to the model's difficulty in synthesizing backgrounds with intricate details when CAM is omitted.

% \begin{figure}[t]
%   \centering
%   \includegraphics[width=0.48\textwidth]{picture/qualitative_exp.pdf}
%   % %%\vspace*{-15pt}
%   \caption{\textbf{Qualitative results of our SCODE approach.}
%   }
%   %%\vspace{-10pt}
%   \label{fig:qulares}
% \end{figure}

% %%\vspace*{-10pt}

% %%\vspace{-7pt}
%-------------------------------------------------------------------------
\subsection{Qualitative Ablation Study}
\label{subsec:abl}
% We conduct an ablation study of our method. As shown in Fig. \ref{fig:ablation}, the (b) column is the result of removing the diffusion module in the camouflage environment generator. Our model without the diffusion process leads to artifacts and fuzzy background compared to our completed model. We think the decrease in synthesis ability is caused by lacking random probabilities. It proves that our diffusion module introduces more probabilities to the background generation, which helps the generator to produce a better environment background.

% To evaluate the effectiveness of our method, we conducted an ablation study. 
% % Specifically, we examined the impact of the diffusion module in the camouflage environment generator. As shown in Fig. \ref{fig:ablation}, the (b) column displays the results obtained after removing the diffusion module from our model.
% As shown in Fig. \ref{fig:ablation}, we observed that without the noise-adding process, our model produced artifacts and a fuzzy background, which are significantly inferior to those obtained by our completed model. While without the camouflage classifier, the model can hardly synthesize the details in the background. We believe that this decrease in synthesis ability is due to the absence of random probabilities introduced by the noise-adding module. Our findings suggest that the noise-adding module and CAM module plays crucial roles in enhancing the generator's ability to produce a better environment background by introducing more probabilities into the background generation process.

To evaluate the effectiveness of our method, we conducted an ablation study. As depicted in Figure~\ref{fig:ablation}, we observed that excluding the noise-adding process resulted in the generation of artifacts and blurry backgrounds, indicating the significance of this module. Similarly, the absence of the camouflage classifier led to inadequate synthesis of background details. We attribute this decline in synthesis quality to the lack of randomness introduced by the noise-adding module. These findings highlight the pivotal roles played by the noise-adding and CAM modules in enhancing the generator's capability to produce realistic environment backgrounds.

\subsection{Quantitative Results}
\label{subsec:quanres}

For quantitative evaluation, we trained the SCODE model on the CAMO and COD10K datasets. The synthetic data generated by the trained SCODE models were then integrated into the training datasets to expand their size. The augmented dataset was subsequently used to train the model alongside the state-of-the-art approach proposed by Fan et al.~\cite{fan2021concealed}. Utilizing four widely recognized evaluation metrics—S-measure ($S_\alpha$), E-measure ($E_\phi$), weighted F-measure ($F_\beta^w$), and Mean Absolute Error ($M$)—our results, detailed in Table~\ref{tab:ModelScore}, demonstrate that our method consistently outperforms other competing techniques by a substantial margin.
% For the quantitative evaluation of our methodology, we train our SCODE model on the datasets (CAMO, COD10k). The synthesized data generated by the trained SCODE models will then be added to the training datasets to expand the original training datasets. We train on the expanded dataset with the current SOTA method~\cite{fan2021concealed}. Note that we do not use the CAMO dataset to train because it only has a testing set.
% Then, we use four evaluation metrics (S-measure($S_\alpha$)~\cite{fan2017structure}, E-measure($E_\phi$)~\cite{Fan2018Enhanced}, weighted F-measure($F_\beta^w$)~\cite{margolin2014evaluate}, Mean Absolute Error($M$)~\cite{perazzi2012saliency}) to assess our model's ability to improve the performance of camouflaged object detection. As shown in Table. \ref{tab:ModelScore}, our method outperforms all other competing methods
% % ~\cite{lin2017feature,he2017mask,zhao2017pyramid,zou2018DLMIA,liu2018picanet,huang2019mask,zhao2019pyramid,wu2019cascaded,chen2019hybrid,le2019anabranch,zhao2019EGNet,fan2020pranet,fan2020camouflaged,fan2021concealed}
% by a notable margin. 
As shown in Table.~\ref{tab:AddScore}, our model achieves a steady improvement in the performance of the three existing models in the plug-and-play setting. 
As shown in Table.~\ref{tab:Comparison_aug}, our SCODE has a better performance than other popular traditional data augmentation methods.
The results show the effectiveness of our proposed approach.
\begin{figure}[h]
  \centering
  \includegraphics[width=.50\textwidth]{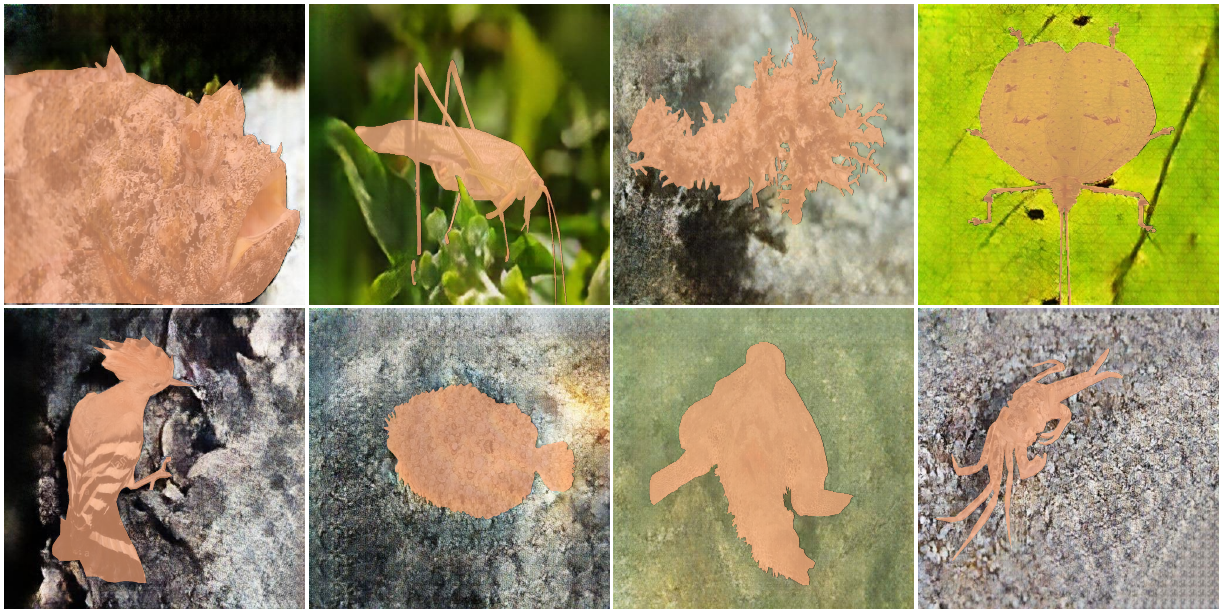}
  % %%\vspace*{-3pt}
  \caption{\textbf{Mask consistency visualization on synthetic data.} These four images are synthesized from our SCODE. The pink areas are the input masks cascaded over the synthesized images.
  }  %%\vspace{-20pt}
  \label{fig:maskconsist}
\end{figure}
% %-------------------------------------------------------------------------
\begin{figure}[h]
  \centering
  \includegraphics[width=.30\textwidth]{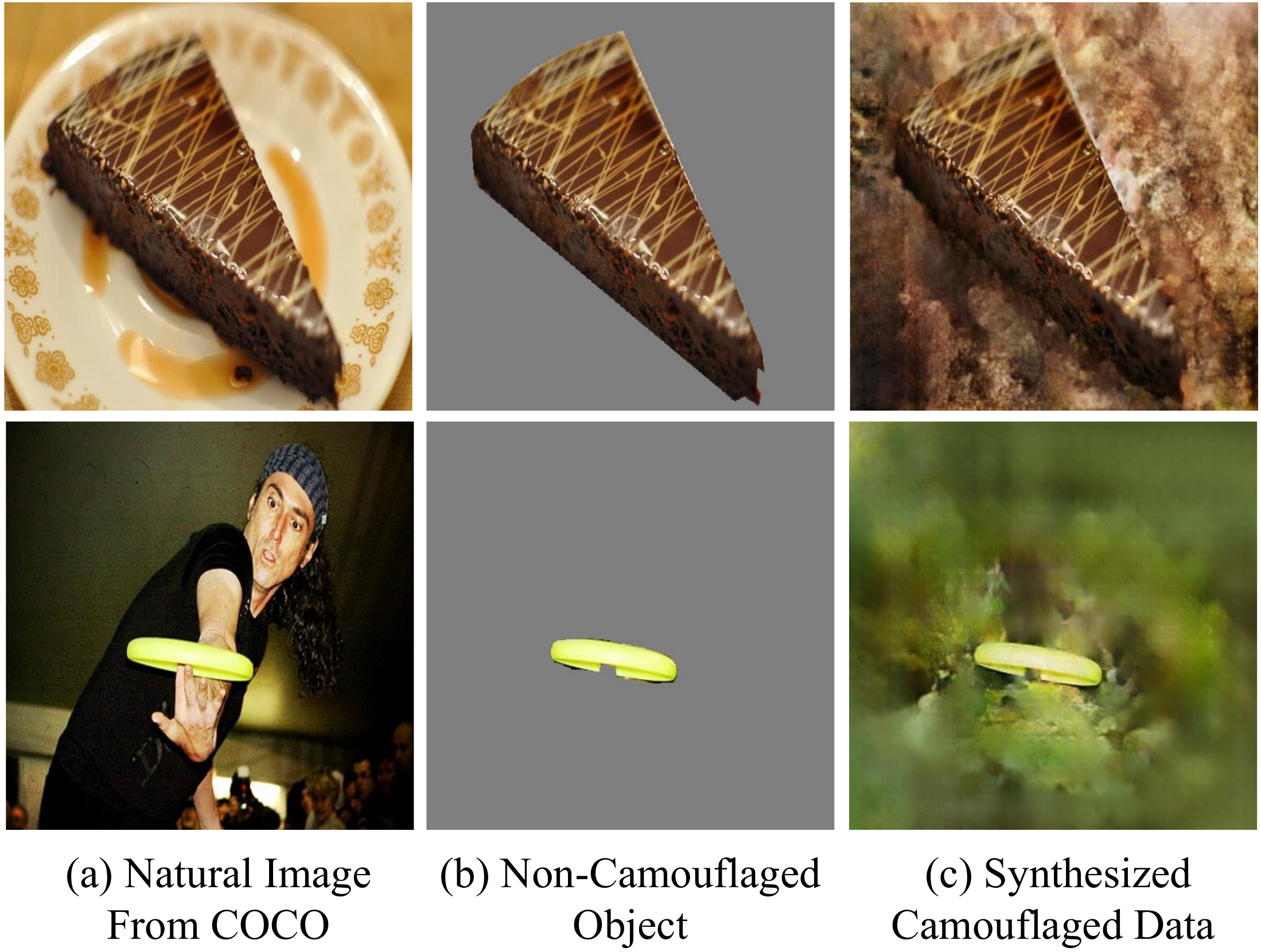}
  % \vspace*{-20pt}
  \caption{The limitation analysis of Test on \textbf{Unseen Out-of-Domain Non-camouflage Input.} Despite that those objects are sampled from a \textbf{unseen} \textit{Non-camouflage} image dataset, its distribution is entirely different from the training dataset. The model still tends to synthesize the corresponding camouflaged backgrounds to help concealment of the input \textit{non-camouflage} object. Please note that the model is \textbf{not trained} on this testing dataset. 
  }  
  \label{fig:outofdist}
\end{figure}
% %-------------------------------------------------------------------------

\subsection{Qualitative Results}
\label{subsec:qualres}
We present a variety of qualitative results generated by the SCODE model in the supplementary materials. To further assess our approach's performance, we compared our qualitative results with those of two previous GAN models, namely CycleGAN~\cite{CycleGAN2017} and pix2pixHD~\cite{wang2018pix2pixHD}. The comparison results, shown in Figure~\ref{fig:qual_compare}, affirm that our model excels in generating high-resolution camouflaged images. This experiment serves as compelling evidence of our approach's proficiency in synthesizing high-quality data for camouflaged object detection.

\label{subsubsec:figure2}
We conducted a comprehensive comparison between our approach and contemporary state-of-the-art generative models, specifically StyleGAN2~\cite{Karras2019stylegan2} and the stable diffusion model~\cite{rombach2021highresolution}. The comparative results are illustrated in Figure~\ref{fig:headcomparison}. 
To enhance the training of the StyleGAN2 model, we initialized the training process with pre-trained parameters from the WildLife dataset~\cite{wildlife} and further fine-tuned the model using the COD10K dataset for a total of 800k iterations. However, the outcomes reveal that StyleGAN2 struggles to effectively capture the distribution of camouflaged images. Consequently, it produces chaotic and unconvincing results when applied to camouflaged object datasets.
Conversely, we employed text prompts to evaluate the stable diffusion model. The prompt texts for each example in the Stable Diffusion Model were as follows: "Camouflaged animal blends in its surroundings," "Green camouflage insects hide in leaves," "Camouflaged fish hiding in the sand on the bottom of the sea, it has the same color as the background to hide from predators," and "Camouflaged white rabbit tries to blend in the snow environment from predators." The results indicate that the stable diffusion model faces challenges in consistently synthesizing camouflaged foreground objects that seamlessly blend with their environments. In some cases, the model generates only the background portion of the text prompt, completely omitting the foreground camouflaged object. Notably, our SCODE model emerges as the sole model capable of effectively synthesizing high-quality camouflaged images, as evidenced in Figure~\ref{fig:headcomparison}. These compelling results further underscore the efficacy of our SCODE approach in generating superior camouflaged images.

\subsection{More Qualitative Results}
\label{supp:subsec:qualres}

We conducted a qualitative evaluation of our SCODE model and present the results in 
% Fig.~\ref{fig:qulares} and 
Fig.~\ref{fig:qulr}. We provide more qualitative results for testing. Since our SCODE model is used as a data augmentation method for the detection method, we use the training set of the COD detection method as the dataset, including both the training and testing sets, for the SCODE model. In the data augmentation process, we perform SCODE on the entire training set of COD, which includes the training and testing sets for SCODE. 

% The results in Fig.~\ref{fig:qulares} show the SCODE training set results, which exhibit a slight difference from the ground truth image. On the other hand, 
Fig.~\ref{fig:qulr} displays the results of the SCODE testing set, where the background is entirely different from the ground truth. We believe that both sets of data augmentation results can contribute to improving the model's ability to detect camouflaged objects effectively. These qualitative results demonstrate that our SCODE model can generate reasonable background images and effectively camouflage the input objects.

% \begin{figure*}[h]
% \centering
%   %\includegraphics[width=0.8\linewidth]{egfigure.eps}
%   \includegraphics[width=0.98\textwidth]{picture/test_imgs.pdf}
%   % \vspace*{-3pt}
%   \caption{More Visualization Results on the test set.}
%    \label{fig:qulr}
% \end{figure*}

\afterpage{%
% \clearpage
\begin{figure*}[p]
\centering
  \includegraphics[width=0.98\textwidth]{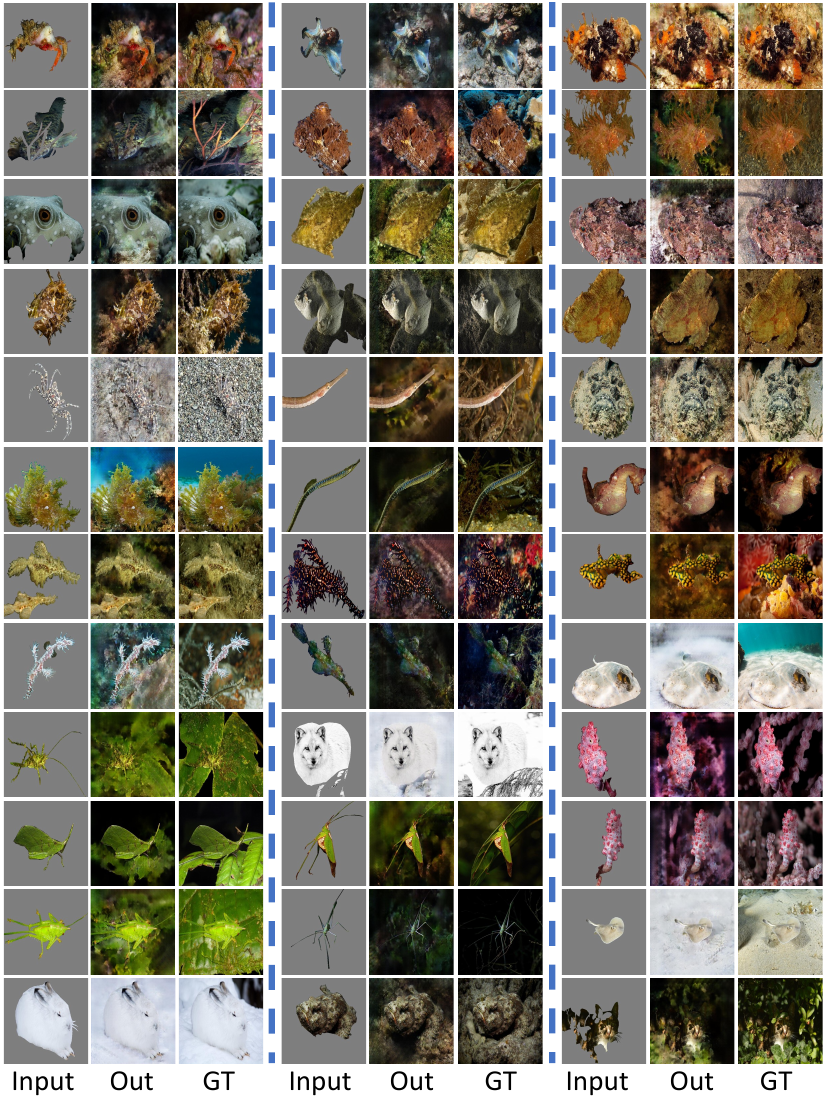}
        \vspace{0.5em} % Add some vertical spacing
    \begin{tabular}{@{}*{9}{>{\centering\arraybackslash}p{.085\textwidth}}@{}}
        Input & Out & GT & Input & Out & GT & Input & Out & GT \\
    \end{tabular}
  \caption{More Visualization Results on the test set.}
   \label{fig:qulr}
\end{figure*}
% \clearpage
}

\subsection{Number of Augmented Data}
Table~\ref{tab:num} provides details about the number of images in the dataset used for training the COD method before and after applying the data augmentation method SCODE. In our quantitative experiments, we generate one augmented data sample for each data point in the training set. The COD model training set is a combination of the CAMO and COD-10k training sets, as the CHAMELEON dataset only contains a testing set. Our data augmentation method utilizes the foreground objects and masks from the COD training set.

%%%%%%%%%%%%%%%%%%%%%%%%%%%%%

\subsubsection{Limitation \& Future Direction - Test on Unseen Out-of-Domain Non-camouflage Dataset}
\label{supp:subsec:ood}
% To conduct a comprehensive analysis of our SCODE model's capabilities, we performed an evaluation on an unseen out-of-domain Non-camouflaged dataset, as depicted in Figure~\ref{fig:outofdist}. Our pre-trained SCODE model, originally trained on camouflage images, was employed without any fine-tuning for this experiment. This task is inherently challenging, given the significant distribution disparity between the COCO-STUFF dataset~\cite{caesar2018coco}, which contains natural scenes, and camouflage images. Despite these challenges, our model manages to produce reasonable results, as demonstrated in the (c) column of Figure~\ref{fig:outofdist}. Although the input objects cannot be effectively camouflaged, our model predicts backgrounds that effectively hide them. 

% Although the current natural-to-camouflage transformation results are still not good enough, this demonstrates the potential of our model for natural-to-camouflage image translation tasks, obviating the need for training on natural datasets. We regard this as a future direction to explore for future camouflaged image synthesis works.

To comprehensively analyze the capabilities of our SCODE model, we conducted an evaluation on an unseen out-of-domain non-camouflaged dataset, as illustrated in Figure2. Our pre-trained SCODE model, originally trained on camouflage images, was utilized without any fine-tuning for this experiment. This task is inherently challenging due to the significant distribution differences between the COCO-STUFF dataset~\cite{caesar2018coco}, containing natural scenes, and camouflage images. Despite these challenges, our model manages to produce reasonable results, as demonstrated in the (c) column of Figure~2. Although the input objects cannot be effectively camouflaged, our model predicts backgrounds that effectively hide them.

While the current natural-to-camouflage transformation results are not optimal, this experiment highlights the potential of our model for natural-to-camouflage image translation tasks without the need for training on natural datasets. We consider this as a promising future direction for camouflaged image synthesis research.

% \bibliographystyle{named}
% \bibliography{ijcai24}

\subsection{Mask Consistency Visualization} 
\label{subsec:mask}
% Fig. \ref{fig:maskconsist} presents the qualitative results of our mask consistency evaluation. Visually, we observe that the mask area remains tightly adhered to the foreground, while the synthesized environment clearly delineates the boundary between the foreground objects and the background. This outcome suggests that our method for generating camouflaged environments effectively preserves the masked object area during the inference process.

Visualized in Figure~\ref{fig:maskconsist}, our mask consistency evaluation demonstrates that the mask area remains well-aligned with the foreground object, while the synthesized environment effectively maintains the boundary between the object and the background. This underscores the success of our method in generating camouflaged environments that accurately preserve the masked object's spatial relationship.

\subsection{Generation Diversity} 

The diversity in image generation is illustrated in Figure~\ref{fig:diversity}. Our SCODE method demonstrates the capability to generate diverse and plausible backgrounds for a single input camouflaged object. This is achieved by skillfully manipulating noise and interpolating the latent space. Notably, our model consistently produces diverse images across both the training and testing datasets, confirming its ability to introduce variability in the generated images.

\subsection{Background Variation on Test Set}
% As shown in Fig. \ref{fig:testvis}, we give more qualitative visualization results of our method to show the diverse generated background of our method. The results Fig. \ref{fig:qulares} show that our SCODE model synthesizes a reasonable background image and camouflages the input objects well. It also shows that the synthesized background has signifciant difference with the ground truth.
As depicted in Fig. \ref{fig:testvis}, we present qualitative visualization results of our method to showcase the diverse backgrounds generated by our SCODE model. Fig. \ref{fig:testvis} demonstrates that our SCODE model effectively synthesizes realistic background images and seamlessly camouflages the input objects. Moreover, it's evident that the synthesized backgrounds significantly differ from the ground truth.

% \begin{figure}[ht]
% \begin{center}
% \includegraphics[width=0.5\textwidth]{rebuttal_fig/StyleFore.jpg}
% \caption*{Diverse Foreground Object Synthesis fed into SCODE}
% \includegraphics[width=0.5\textwidth]{rebuttal_fig/StyleWhole.jpg}
% \caption*{Stylegan can't synthesize entire camouflage image}
% \end{center}
% \caption{Experiments on camouflaged foreground object synthesis  }  
% \label{fig:fore_gen}
% \end{figure}

\subsection{Camouflaged Foreground Object Synthesis}
To explore diverse foreground camouflage object generation, we employed StyleGAN to synthesize foreground elements, which were subsequently fed into SCODE. As shown in Fig.~\ref{fig:fore_gen}, our experiments with StyleGAN successfully generated a variety of camouflaged foreground objects with visually persuasive results. Despite StyleGAN's limitations in crafting entire camouflaged scenes, its efficacy in producing diverse foreground elements is undeniable. Leveraging these diverse foregrounds, our SCODE framework can effectively synthesize complete images, enriching the camouflage with varied foreground elements.
\section{Conclusion}
\label{sec:cls}
%%\vspace{-5pt}
% We introduce the idea of using GAN-synthesized data to solve the scarcity of data quantity problem in camouflaged object datasets. 
% We first introduce a plug-and-play module to expand the image dataset for camouflaged object detection. Secondly, we propose a camouflage environment generator, which includes the diffusion module and camouflage distribution loss.
% Thirdly, we evaluate the effectiveness of our proposed method on three camouflage datasets through camouflaged object detection tasks. The results show that our method outperforms all other competing methods by a notable margin.
% We also conduct experiments to see the synthesis ability and show our method's potential to directly transfer natural images into camouflage images without training. 
In conclusion, our work provides a new solution for the camouflaged object detection method to solve the data quantity and quality problem and increase camouflage segmentation performance.
In this work, we presented a novel approach to address the scarcity of data in camouflaged object datasets by using GAN-synthesized data. Our proposed method consists of a plug-and-play module that can expand the dataset for camouflaged object detection, a camouflage environment generator that includes the noise-adding module and camouflage distribution loss, and an evaluation of our method on three different camouflage datasets through camouflaged object detection tasks. The experiments demonstrate that our approach outperforms all other competing methods by a significant margin, and our method's potential to transfer out-of-domain natural images into camouflage images without training was also explored. In conclusion, our work provides a new solution to improve the performance of camouflaged object detection by addressing the data scarcity problem.

% In this work, we introduced a novel approach to address the data scarcity challenge in camouflaged object datasets by leveraging GAN-synthesized data. Our method encompasses a plug-and-play module for dataset expansion, a camouflage environment generator comprising the noise-adding module and camouflage distribution loss, and an extensive evaluation across multiple camouflage datasets through camouflaged object detection tasks. Our experiments demonstrated the substantial superiority of our approach over existing methods, and we explored the potential of our model for natural-to-camouflage image translation tasks. In summary, our work provides a promising solution to enhance the performance of camouflaged object detection by effectively mitigating the data scarcity issue.

% \bibliography{aaai24}

% \end{document}

\appendices
\ifCLASSOPTIONcaptionsoff
  \newpage
\fi



%\bibliographystyle{IEEEtran}

%%%%%%%%% REFERENCES
{\small
\bibliographystyle{ieee_fullname}
\bibliography{main}
}

\end{document}